\documentclass[conference]{IEEEtran}
\IEEEoverridecommandlockouts
\usepackage{cite}
\usepackage{amsmath,amssymb,amsfonts,amsthm}
\usepackage{algorithmic}
\usepackage{graphicx}
\usepackage{textcomp}
\usepackage{xcolor}
\usepackage{booktabs}
\usepackage{multirow}
\usepackage{algorithm}
\usepackage{tikz}
\usepackage{float}
\usetikzlibrary{arrows.meta}
 
\newcommand{\ours}{CDN}
\newtheorem{proposition}{Proposition}
\newtheorem{lemma}{Lemma}
\newtheorem{remark}{Remark}

\newcommand{\R}{\mathbb{R}}
\newcommand{\bx}{\mathbf{x}}
\newcommand{\by}{\mathbf{y}}
\newcommand{\bt}{\mathbf{t}}
\newcommand{\bm}{\mathbf{m}}
\newcommand{\bb}{\mathbf{b}}
\newcommand{\bW}{\mathbf{W}}

\newcommand{\bu}{\mathbf{u}}
\newcommand{\cN}{\mathcal{N}}
\newcommand{\cX}{\mathcal{X}}

\def\BibTeX{{\rm B\kern-.05em{\sc i\kern-.025em b}\kern-.08em
    T\kern-.1667em\lower.7ex\hbox{E}\kern-.125emX}}
\begin{document}

\title{Anomalies as Residuals: Cauchy Displacement Networks}

\author{
\IEEEauthorblockN{Yang Cao}
\IEEEauthorblockA{
\textit{School of Computing and Information Technology,}\\
\textit{Great Bay University;}\\
\textit{Tsinghua Shenzhen International Graduate School,}\\
\textit{Tsinghua University}\\
China\\
charles.cao@ieee.org}
\and
\IEEEauthorblockN{Zihao Guo*}
\IEEEauthorblockA{\textit{Institute for Advanced Research,}\\
\textit{Great Bay University;}\\
% \textit{School of Mathematical Sciences}\\
% \textit{University of Science and Technology of China}\\
China \\
guozh@gbu.edu.cn}
\thanks{*Corresponding author.}
}

\maketitle

% =================================================================
\begin{abstract}
Neighbourhood-based detectors such as KNN, LOF and LUNAR remain among the most accurate methods for tabular anomaly detection, but they compare every query with the stored training set, so their inference cost and memory grow with the data and the data must be kept at deployment. We ask whether the neighbourhood geometry these detectors exploit can be compiled into a small network that scores a query with a single forward pass. We propose the Cauchy Displacement Network (CDN), which regresses each normal point onto the mean of its $k$ nearest training neighbours with a single-hidden-layer network of Cauchy activations and scores a query by the length of its predicted displacement. This displacement is an empirical mean-shift vector, whose fixed-radius population counterpart locally approximates a scaled gradient of the log density. We show that Cauchy units exactly represent the score fields of a structured family of Student-$t$ expert densities, and establish a constructive approximation bound for the corresponding population neighbourhood-mean map using a single-hidden-layer Cauchy network. We further show that the bounded network output guarantees asymptotically linear residual-score growth along every fixed direction. On 46 benchmark datasets against 28 baselines, CDN obtains the best mean rank under both AUC-ROC and AUC-PR, while scoring $10^4$ queries in a few milliseconds on a single CPU thread, up to three orders of magnitude faster than neighbourhood-based detectors, with a model whose size does not depend on the training set.

\end{abstract}


% =================================================================
\section{Introduction}

Anomaly detection identifies observations that deviate from the dominant patterns of a dataset and supports applications such as fraud detection, intrusion detection, log monitoring and industrial quality control \cite{chandola2009anomaly,pang2021deep,liu2024deep,song2023adops,tang2024substructure}. In many of these applications a detector is trained once on historical normal data and then deployed to score new records over a long period, often on devices with limited computation or in settings where the training data cannot be retained. The cost that matters at deployment is therefore the cost of scoring a query, together with the memory and data that must be kept to compute that score.

On tabular data, neighbourhood methods such as KNN \cite{ramaswamy2000efficient}, LOF \cite{breunig2000lof} and LUNAR \cite{goodge2022lunar} are strong baselines \cite{han2022adbench}. These methods require access to reference data at inference time. Under brute-force search, scoring a query involves $O(n)$ distance evaluations and $O(nd)$ arithmetic operations. Reconstruction and one-class networks \cite{aggarwal2016,kingma2013auto,zong2018deep,ruff2018deep} avoid this dependency but can lag behind neighbourhood methods, while recent diffusion and flow-matching detectors \cite{livernochediffusion,li2025scalable} introduce more expensive models. This creates an accuracy--deployment trade-off.

In data-intensive monitoring systems, historical records may be maintained in a central repository, while anomaly scoring is performed by services with limited memory or strict latency requirements. Reference-based detectors must either access these records at query time or maintain a searchable copy of them. Our goal is to remove this deployment dependency while retaining the local geometric information used by neighbourhood detectors.

%On tabular data, the most accurate detectors are still largely neighbourhood-based, KNN \cite{ramaswamy2000efficient} and LOF \cite{breunig2000lof} remain strong on standard benchmarks \cite{han2022adbench}, and LUNAR \cite{goodge2022lunar} improves them by learning a graph neural network over nearest-neighbour distances. Their accuracy comes at a structural price: at test time every query must be compared with the stored training set, which costs $O(nd)$ distance evaluations for $n$ training points in $d$ dimensions, or an index whose efficiency degrades in high dimensions, and the full training set must be kept at deployment. Detectors that do not need the training data at test time, such as reconstruction-based \cite{aggarwal2016,kingma2013auto,zong2018deep} and one-class \cite{ruff2018deep} networks, are often less accurate than these simple neighbourhood methods, while recent generative detectors based on diffusion or flow matching \cite{livernochediffusion,li2025scalable} rely on large networks and long training. Accuracy and deployability thus pull in opposite directions.

This raises the question of whether the neighbourhood geometry exploited by these detectors can be encoded into a compact
parametric model, allowing queries to be scored without retaining the training data. A natural alternative is self-reconstruction, where a network is trained to reproduce normal inputs. However, the reconstruction objective does not explicitly enforce neighbourhood geometry. When the hypothesis class is sufficiently expressive, an identity-like mapping may achieve a small training loss while also reconstructing unseen anomalies accurately. Although bottlenecks and regularisation can mitigate this effect, they do not directly impose the local geometric supervision provided by neighbourhood statistics.

We propose the Cauchy Displacement Network (\ours), which combines neighbourhood-mean regression with a compact single-hidden-layer Cauchy architecture. During training, each normal point is mapped to the mean of its $k$ nearest training neighbours rather than to itself. The identity map generally has nonzero loss for this target, encouraging the network to learn local neighbourhood geometry. At test time a query is scored by the length of the displacement between the query and the network output, which requires one forward pass of a single-hidden-layer network and no neighbour search. The network uses the Cauchy activation of XNet \cite{li2025cauchy}, a rational function derived from the Cauchy integral formula. We refer to the discrepancy between a query and its predicted neighbourhood mean as its residual. Normal observations are expected to produce smaller residuals, whereas anomalous observations may produce larger ones.

This design has a direct statistical interpretation. The displacement from a point to the mean of its neighbours is the mean-shift vector, which is approximately proportional to the gradient of the log density \cite{fukunaga1975estimation,comaniciu2002mean}, that is, the score function of the data distribution. \ours\ therefore learns a neighbourhood-mean mapping whose displacement is locally related to the score field, and uses the magnitude of the predicted displacement as its anomaly score. We further show that 
the Cauchy activation is particularly relevant to this setting because its rational form exactly represents the score fields of a structured family of Student-$t$ expert \cite{welling2003learning} densities. We further establish a constructive approximation bound for the corresponding population neighbourhood-mean map. In addition, the bounded output of the direct Cauchy network ensures asymptotically growing residual scores for sufficiently distant queries. This latter property is shared by other bounded-activation architectures, whereas the exact score representation exploits the specific rational structure of Cauchy units.

% The main contributions of this work are as follows.
% \begin{itemize}
% \item We propose \ours, a neighbourhood-based regression detector that replaces self-reconstruction with neighbour-mean regression. It requires no negative sampling, data augmentation or generative modelling, and it scores a query with one forward pass, without storing the training data or searching for neighbours.
% \item We provide a theoretical foundation for \ours: the population counterparts of its neighbourhood-mean displacements locally approximate a scaled gradient of the log density; Cauchy units exactly represent the score fields of a structured family of Student-$t$ expert densities; and a single-hidden-layer Cauchy network admits a constructive approximation error bound for the corresponding population neighbourhood-mean map. We further establish that this approximation preserves anomaly-score separation when the population score has a positive margin exceeding twice the approximation error. Finally, the bounded network output guarantees asymptotically linear residual-score growth along every direction.
% \item Experiments on 46 benchmark datasets against 28 baselines show that \ours\ obtains the best mean rank under both AUC-ROC and AUC-PR, ahead of KNN, MOD and LUNAR, while scoring $10^4$ queries in a few milliseconds on a single CPU thread, up to three orders of magnitude faster than neighbourhood-based detectors, with a model whose size does not depend on the training set.
% \end{itemize}

The main contributions of this work are to:
\begin{itemize}
\item Propose \ours, which regresses each normal point onto the mean of its $k$ nearest training neighbours with a single-hidden-layer Cauchy network and scores a query by the length of its displacement in one forward pass, without storing the training data or searching for neighbours.
\item Establish a statistical interpretation of neighbourhood-mean regression through a fixed-radius population mean-shift expansion, and show that Cauchy units exactly represent the density scores of products of Student-$t$ experts.

\item Derive a constructive approximation bound for the corresponding population neighbourhood-mean map, establish conditional preservation of anomaly-score margins, and prove asymptotically linear far-field growth of the residual score.

\item Evaluate \ours\ on 46 benchmark datasets against 28 baselines, where it obtains the best mean rank under both AUC-ROC and AUC-PR while scoring queries up to three orders of magnitude faster than neighbourhood-based detectors.
\end{itemize}

% =================================================================
\section{Related Work}

\subsection{Classical Detectors}
Distance-based detectors use nearest-neighbour distances \cite{ramaswamy2000efficient,angiulli2002fast}, sometimes approximated by Sampling \cite{sugiyama2013rapid}. Density-based methods include LOF \cite{breunig2000lof}, its variants \cite{tang2002enhancing,papadimitriou2003loci}, and FeatureBagging \cite{lazarevic2005feature}; ABOD uses neighbour angles \cite{kriegel2008angle}. Statistical alternatives include HBOS \cite{goldstein2012histogram}, COPOD \cite{li2020copod}, ECOD \cite{li2022ecod}, GMM \cite{aggarwal2016}, PCA \cite{shyu2003novel}, and QMCD \cite{fang2001wrap}. Isolation methods extend iForest \cite{liu2008isolation,liu2012isolation} through oblique splits \cite{hariri2019extended}, hashing \cite{zhang2017lshiforest}, hyperspheres \cite{bandaragoda2018isolation,duranianomaly} or distributional kernels \cite{ting2020isolation,ting2021isolation,cao2025anomaly}. Unlike reference-based scorers, \ours\ needs only trained parameters and standardisation statistics.

\subsection{Deep Detectors}

Reconstruction-based detectors use AE \cite{aggarwal2016}, VAE \cite{kingma2013auto} or DAGMM \cite{zong2018deep}; one-class methods include OCSVM \cite{scholkopf1999support}, SVDD \cite{tax2004support} and DeepSVDD \cite{ruff2018deep}. Adversarial approaches include AnoGAN \cite{schlegl2019f}, ALAD \cite{zenati2018adversarially}, GANomaly \cite{akcay2018ganomaly} and SO-GAAL \cite{liu2019generative}. Self-supervised methods include GOAD \cite{bergmanclassification}, ICL \cite{shenkar2022anomaly}, SLAD \cite{xu2023fascinating} and DIF \cite{xu2023deep}. LUNAR \cite{goodge2022lunar} learns over neighbour distances but still searches at inference. Diffusion and flow-based methods \cite{livernochediffusion,li2025scalable} use more elaborate generative training. \ours\ instead combines geometric supervision with single-pass, data-free inference.

%Reconstruction-based detectors train an autoencoder \cite{aggarwal2016} or a variational autoencoder \cite{kingma2013auto} on normal data and use reconstruction error as the score, and DAGMM \cite{zong2018deep} combines the latent code and reconstruction features with a Gaussian mixture. One-class methods map normal data into a compact region, as in OCSVM \cite{scholkopf1999support}, SVDD \cite{tax2004support} and DeepSVDD \cite{ruff2018deep}. Adversarial detectors include AnoGAN \cite{schlegl2019f}, ALAD \cite{zenati2018adversarially}, GANomaly \cite{akcay2018ganomaly} and SO-GAAL \cite{liu2019generative}. Self-supervised detectors predict random transformations (GOAD \cite{bergmanclassification}), contrast internal feature partitions (ICL \cite{shenkar2022anomaly}), learn from subspaces of different scales (SLAD \cite{xu2023fascinating}) or apply isolation to random representations (DIF \cite{xu2023deep}). LUNAR \cite{goodge2022lunar} learns a graph neural network over nearest-neighbour distances and requires a neighbour query at test time. Diffusion-based detectors use denoising reconstruction or estimated diffusion time \cite{livernochediffusion}, and TCCM \cite{li2025scalable} learns a time-conditioned contraction field with flow matching. \ours\ differs from reconstruction methods in its target, from LUNAR in requiring no neighbour search at test time, and from diffusion and flow models in its much smaller network and lower training cost.

\subsection{Mean Shift and Score Functions}
Mean shift moves a point to the weighted mean of its neighbours and was introduced as an estimator of the density gradient \cite{fukunaga1975estimation}; it is widely used for mode seeking and clustering \cite{comaniciu2002mean}. The mean-shift outlier detector (MOD) \cite{yang2021mean} shifts each point to the mean of its $k$ nearest neighbours for a few iterations and uses the distance between the original and the shifted position as the outlier score. Such methods compute the shift nonparametrically and therefore require a neighbour search over the stored data for every query, repeated at each iteration. In representation learning, denoising autoencoders were shown to learn the score function of a smoothed data density \cite{vincent2011connection}. \ours\ connects neighbourhood-based anomaly detection with parametric regression. It computes one-step nearest-neighbour means for normal training samples and uses them as supervised targets for a compact network. Unlike MOD, which evaluates iterative nonparametric shifts for each query, \ours\ predicts the neighbourhood mean through a single forward pass without test-time neighbour searches.

\subsection{Cauchy Activation and XNet}
XNet \cite{li2025cauchy} introduces the Cauchy activation, a rational function obtained by discretising the Cauchy integral formula, and shows that networks built from it approximate functions with sharp local features efficiently. A single-hidden-layer XNet achieves fast approximation rates and compares favourably with Kolmogorov-Arnold networks \cite{li2026kolmogorov}, and Cauchy-type architectures have been applied to compact, data-efficient learning \cite{zhang2025cauchynet} and to high-dimensional partial differential equations \cite{zheng2026xnet}. In this work, the Cauchy activation is motivated not only by its approximation capability but also by its structural compatibility with density-score fields. In particular, its rational components match the logarithmic derivatives of Student-$t$-type expert factors, enabling the exact representation established
in Proposition~\ref{prop:pot}.

% =================================================================
\section{Problem Formulation}

Let $\cX=\{\bx_1,\dots,\bx_n\}\subset\R^d$ be a training set drawn from the distribution of normal data with density $p$. A detector learns a scoring function $s:\R^d\to\R$ from $\cX$ and assigns higher scores to anomalies. At test time it receives queries that may be normal or anomalous and is evaluated by how well $s$ ranks anomalies above normal queries. This is the semi-supervised, or one-class, setting adopted in recent tabular anomaly detection benchmarks \cite{han2022adbench,livernochediffusion}. Beyond accuracy, we are interested in the cost of scoring a query and in the state that must be kept after training. We call a detector \emph{data-free at inference} if $s$ can be evaluated without access to $\cX$.

Throughout the paper, $\|\cdot\|$ denotes the Euclidean norm, $\cN_k(\bx)$ denotes the indices of the $k$ nearest neighbours of $\bx$ in $\cX$ (excluding $\bx$ itself when $\bx\in\cX$), and $r_k(\bx)$ is the distance from $\bx$ to its $k$-th nearest neighbour.

% =================================================================
\section{Cauchy Displacement Networks}
\label{sec:method}

\begin{figure*}[!t]
\centering
\begin{tikzpicture}[
  font=\footnotesize,
  box/.style={draw, rounded corners=2pt, align=center, minimum height=0.8cm, inner sep=2.5pt},
  nd/.style={circle, draw, minimum size=0.38cm, inner sep=0pt},
  cn/.style={circle, draw, fill=red!12, minimum size=0.5cm, inner sep=0pt},
  arr/.style={-{Stealth[length=2mm]}, thick},
  lab/.style={font=\scriptsize\itshape, text=black!70}
]
\node[font=\footnotesize\bfseries] at (4.3,1.25) {Training};
\node[font=\footnotesize\bfseries] at (13.2,1.25) {Inference};
% data pipeline
\node[box, fill=blue!6, minimum width=2.3cm] (data) at (0,0) {Normal training\\data $\cX$};
\node[box, fill=blue!6, minimum width=2.3cm] (knn) at (0,-1.3) {$k$-NN search\\(once, offline)};
\node[box, fill=blue!6, minimum width=2.3cm] (tgt) at (0,-2.75) {Targets\\$\bt_i=\frac{1}{k}\sum_{j\in\cN_k(\bx_i)}\bx_j$};
\draw[arr] (data) -- (knn);
\draw[arr] (knn) -- (tgt);
% network
\foreach \i/\y in {1/0,2/-0.6,3/-1.2,4/-1.8} \node[nd] (x\i) at (3.6,\y) {};
\foreach \j/\y in {1/0.3,2/-0.35,3/-1.0,4/-1.65,5/-2.3} \node[cn] (h\j) at (5.2,\y) {\tiny$\phi_{\j}$};
\foreach \i/\y in {1/0,2/-0.6,3/-1.2,4/-1.8} \node[nd] (o\i) at (6.8,\y) {};
\foreach \i in {1,...,4} \foreach \j in {1,...,5} \draw[black!30] (x\i) -- (h\j);
\foreach \j in {1,...,5} \foreach \i in {1,...,4} \draw[black!30] (h\j) -- (o\i);
\node at (3.6,-2.35) {$\bx_i$};
\node at (6.8,-2.35) {$f_\theta(\bx_i)$};
\node[lab] at (4.35,0.62) {$\bW_1,\bb_1$};
\node[lab] at (6.05,0.62) {$\bW_2,\bb_2$};
\draw[arr] (data.east) -- (3.38,0);
% activation inset
\draw[black!40, rounded corners=2pt] (4.05,-2.72) rectangle (6.35,-4.5);
\node[lab] at (5.2,-2.93) {Cauchy unit $j$};
\draw[->, black!50] (4.2,-3.6) -- (6.2,-3.6);
\draw[->, black!50] (5.2,-3.98) -- (5.2,-3.1);
\draw[thick, red!70!black, domain=-2.8:2.8, samples=100] plot ({5.2+0.33*\x}, {-3.65+0.4*(0.4*\x+0.25)/(\x*\x+0.36)});
\node[font=\scriptsize] at (5.2,-4.18) {$\phi_j(z)=\frac{\alpha_j z+\beta_j}{z^2+\gamma_j^2+\epsilon}$};
% loss
\node[box, fill=orange!10, minimum width=2.4cm] (loss) at (8.75,-0.9) {Loss\\$\frac{1}{nd}\sum_i\|f_\theta(\bx_i)-\bt_i\|^2$};
\draw[arr] (7.05,-0.9) -- (loss.west);
\draw[arr] (tgt.south) -- (0,-4.75) -| node[pos=0.2, above, lab]{$\bt_i$} (loss.south);
\draw[arr, dashed] (loss.north) |- node[pos=0.75, above, lab]{update $\theta$ (Adam, $T$ steps)} (5.2,0.95) -- (5.2,0.57);
% separator
\draw[black!40, dashed] (10.4,1.4) -- (10.4,-5.0);
% inference
\node[box, fill=green!8, minimum width=2.2cm] (q) at (12.0,0) {Query $\bx$};
\node[box, fill=red!8, minimum width=2.2cm] (net) at (12.0,-1.4) {CDN\\one forward pass};
\node[box, fill=green!8, minimum width=2.2cm] (out) at (12.0,-2.8) {$f_\theta(\bx)$};
\node[box, fill=yellow!15, minimum width=2.6cm] (sc) at (14.8,-2.8) {Score\\$s(\bx)=\|\bx-f_\theta(\bx)\|$};
\draw[arr] (q) -- (net); \draw[arr] (net) -- (out); \draw[arr] (out) -- (sc);
\draw[arr] (q.east) -| (sc.north);
\node[lab, align=center] at (13.4,-4.05) {No training data stored,\\no neighbour search};
\end{tikzpicture}
\caption{Overview of \ours. Training: each normal point is paired with the mean of its $k$ nearest training neighbours, and a single-hidden-layer network with Cauchy activations regresses the point onto this target. Inference: a query is scored by the length of the displacement between the query and the network output, using only the trained parameters.}
\label{fig:framework}
\end{figure*}

Fig.~\ref{fig:framework} gives an overview of \ours. Training has two stages. The first stage computes, once and offline, a regression target for each normal point: the mean of its $k$ nearest training neighbours. The second stage trains a single-hidden-layer network with Cauchy activations to map each point to its target. At inference, the network maps a query to a predicted neighbour mean, and the anomaly score is the distance between the query and this prediction. The training data and the neighbour index are discarded after training. This section motivates the design (Section~\ref{sec:method-motivation}) and then describes the targets, the network, the anomaly score, the training procedure and the complexity.

\subsection{Motivation: Beyond Self-Reconstruction}
\label{sec:method-motivation}

A reconstruction-based detector, such as an autoencoder (AE), trains a network $g$ to minimise $\frac{1}{n}\sum_i\|g(\bx_i)-\bx_i\|^2$ and scores a query by $\|\bx-g(\bx)\|$. The identity map attains the minimum value zero. Whenever the hypothesis class can represent or closely approximate the identity, the reconstruction objective admits an identity-like solution with a small training loss. Consequently, minimising this objective alone does not necessarily encourage the model to learn anomaly-sensitive structure. Such a solution may also reconstruct unseen anomalies accurately. This happens readily on tabular data. A single hidden layer of width $h\ge 2d$ represents the identity exactly with ReLU units, since $x=\max(x,0)-\max(-x,0)$, Cauchy units approximate it on any bounded region (Lemma~\ref{lem:id}), and benchmark datasets often have only a few to a few dozen features, so a hidden width of 64 or 128 is wider than the input on most of them. Bottleneck architectures and regularisation can mitigate this shortcut, although their effectiveness depends on the architecture and data distribution. Our motivation concerns the absence of explicit neighbourhood supervision in self-reconstruction, rather than a universal failure of autoencoders.

\begin{figure}[!t]
\centering
\includegraphics[width=0.48\textwidth]{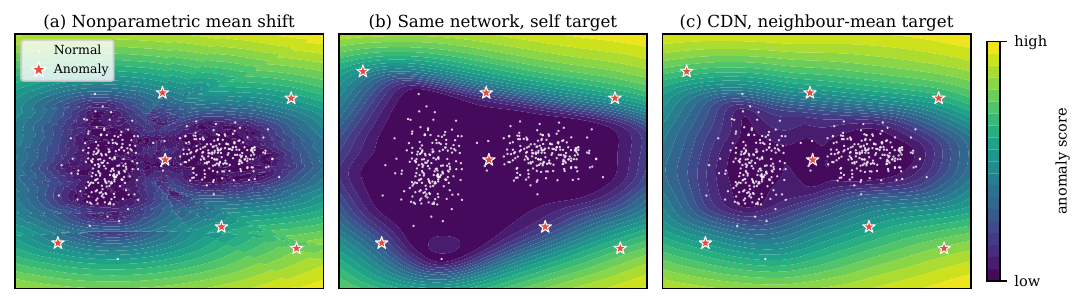}
\caption{Anomaly scores on a two-dimensional example. White dots are normal training points from two clusters; red stars are example anomalies. The colour at each location is the score a query there would receive, rescaled to $[0,1]$ within each panel (dark: low, judged normal; bright: high, judged anomalous). (a) Nonparametric mean shift $\|\bx-\bt(\bx)\|$, which needs the training data and a neighbour search for every query. (b) The same Cauchy network trained to reconstruct its input: the low-score region covers both clusters and a large empty area around them. (c) \ours: the low-score region follows the clusters as in (a), using only the network. All networks use the configuration of Section~\ref{sec:method-training}.}
\label{fig:toy}
\end{figure}

Fig.~\ref{fig:toy}(b) illustrates the problem on a two-dimensional example with two clusters. A network trained to reconstruct its input assigns low scores to a broad region that contains both clusters, the gap between them and a wide margin around them. The network has learned an approximation of the identity on this region rather than the shape of the normal data.

\subsection{Neighbour-Mean Targets}
\label{sec:method-target}

\ours\ keeps the reconstruction architecture but replaces the target. For each training point $\bx_i$ we define
\begin{equation}
\bt_i=\frac{1}{k}\sum_{j\in\cN_k(\bx_i)}\bx_j ,\qquad i=1,\dots,n,
\label{eq:target}
\end{equation}
where $\cN_k(\bx_i)$ excludes $\bx_i$ itself. The difference $\bm_i=\bt_i-\bx_i$ is the $k$-nearest-neighbour mean-shift vector of $\bx_i$ \cite{fukunaga1975estimation,comaniciu2002mean}. When a point lies in a locally balanced neighbourhood, the neighbouring observations may approximately cancel in the mean-shift vector, leading to a small displacement. Near the boundary of a cluster, asymmetric neighbourhoods can produce a displacement towards denser regions. The displacement magnitude depends on both neighbour distances and directions, and is not determined by the neighbourhood radius alone.

\begin{table}[!t]
\centering
\caption{Neighbour-mean targets on four one-dimensional points with $k=2$. $r_2$ is the distance to the second nearest neighbour, i.e.\ the score of the KNN detector.}
\label{tab:example}
\small
\begin{tabular}{cccccc}
\toprule
$x_i$ & $\cN_2(x_i)$ & $t_i$ & $m_i=t_i-x_i$ & $|m_i|$ & $r_2(x_i)$\\
\midrule
0 & $\{1,2\}$ & 1.5 & $\phantom{-}1.5$ & 1.5 & 2\\
1 & $\{0,2\}$ & 1.0 & $\phantom{-}0.0$ & 0.0 & 1\\
2 & $\{1,0\}$ & 0.5 & $-1.5$ & 1.5 & 2\\
5 & $\{2,1\}$ & 1.5 & $-3.5$ & 3.5 & 4\\
\bottomrule
\end{tabular}
\end{table}

Table~\ref{tab:example} gives a small example. The isolated point $5$ has the longest shift. The point $1$ has one neighbour on each side, their mean coincides with the point, and its shift vanishes, although its neighbour distance $r_2=1$ is half that of the points $0$ and $2$. The shift therefore separates the interior of the data from its fringe more sharply than the neighbour distance does, and its sign points towards the data: positive at the left end, negative at the right. \ours\ is trained to reproduce the column $t_i$ and scores a point by the column $|m_i|$.

Neighbourhood-mean regression changes the role of the identity map. Let $\mathcal{L}(f)=\frac{1}{nd}\sum_i \|f(\bx_i)-\bt_i\|^2$.
For the identity map,
\begin{equation}
\mathcal{L}(\mathrm{id})
=\frac{1}{nd}\sum_{i=1}^{n}\|\bm_i\|^2 .
\end{equation}
Thus, $\mathcal{L}(\mathrm{id})>0$ whenever at least one training point has a nonzero mean-shift displacement. Unlike self-reconstruction, the proposed targets explicitly encode the directions and magnitudes of local neighbourhood shifts. Minimising this objective encourages the network to learn such displacement information from normal training samples. However, accurate fitting of these targets and generalisation to unseen queries depend on the network capacity and optimisation, and are not guaranteed by the objective alone.

Two design choices deserve comment. First, the targets use the plain neighbour mean, without distance weights or iteration as in mean-shift clustering \cite{comaniciu2002mean} or MOD \cite{yang2021mean}. We adopt a single step to obtain a simple,
fixed regression target without iterative mode seeking. Whether one step provides sufficient anomaly-discriminative information
depends on the data distribution and is evaluated empirically. Second, the targets are fixed before training and computed with exact search. Their cost is paid once, and the size of the neighbourhood $k$ is the only parameter they introduce.

\subsection{Network Architecture and Cauchy Activation}
\label{sec:method-network}

We adopt a single-hidden-layer XNet \cite{li2025cauchy} to learn the neighbourhood-mean mapping:
\begin{equation}
f_\theta(\bx)
=
\bW_2\phi\!\left(\bW_1\bx+\bb_1\right)+\bb_2,
\label{eq:net}
\end{equation}
where $\bW_1\in\R^{h\times d}$, $\bb_1\in\R^h$, $\bW_2\in\R^{d\times h}$, $\bb_2\in\R^d$, and $h$ denotes the hidden width.

The Cauchy activation is motivated by the Cauchy integral formula for analytic functions,
\begin{equation}
F(z)=\frac{1}{2\pi i}
\oint_{\partial D}
\frac{F(\zeta)}{\zeta-z}\,d\zeta.
\end{equation}
For real $x$ and $\zeta=a+i\gamma$ with $\gamma>0$, the Cauchy kernel satisfies
\begin{equation}
\frac{1}{\zeta-x}
=
\frac{a-x-i\gamma}{(x-a)^2+\gamma^2}.
\end{equation}
Up to constant factors, its real and imaginary parts are $z/(z^2+\gamma^2)$ and $1/(z^2+\gamma^2)$ with $z=x-a$, which motivate these rational basis functions. Their trainable linear combination yields the Cauchy activation \cite{li2025cauchy}
\begin{equation}
\phi_j(z)
=
\frac{\alpha_j z+\beta_j}
{z^2+\gamma_j^2+\epsilon},
\label{eq:cauchy}
\end{equation}
where $(\alpha_j,\beta_j,\gamma_j)$ are trainable parameters for each hidden unit, and $\epsilon>0$ is a small constant for numerical stability.

The activation contains two complementary components. Writing $\widetilde{\gamma}_j=\sqrt{\gamma_j^2+\epsilon}$, the odd component $\alpha_j z/(z^2+\widetilde{\gamma}_j^2)$ changes sign at $z=0$, attains extrema at $z=\pm\widetilde{\gamma}_j$, and decays as $O(|z|^{-1})$ for large $|z|$. It therefore provides a sign-changing response along the projection coordinate $\mathbf w_j^\top\bx+b_{1,j}$, with algebraically decaying tails away
from the corresponding hyperplane. The even component $\beta_j/(z^2+\widetilde{\gamma}_j^2)$ provides a localised response centred on the same hyperplane, with amplitude $\beta_j/\widetilde{\gamma}_j^2$ and algebraically decaying tails. The trainable scale parameter allows different units to respond to structures at different scales.

The population mean-shift displacement is locally related to the gradient of the log density, and the odd rational component of the Cauchy activation matches the logarithmic derivatives of Student-$t$ expert factors. As shown in Proposition~\ref{prop:pot}, a single hidden layer of Cauchy units can represent the score field of a structured family of Student-$t$ expert densities exactly. Moreover, the boundedness of the Cauchy activations ensures that the network output remains within a bounded distance of $\bb_2$, yielding the far-field anomaly-score guarantee in Proposition~\ref{prop:far}.

We use the direct form~\eqref{eq:net} rather than $f_\theta(\bx)=\bx+g_\theta(\bx)$. A bounded skip-connected displacement $g_\theta$ yields a bounded residual score $\|g_\theta(\bx)\|$, whereas a bounded-output direct network produces diverging residuals as $\|\bx\|\to\infty$ (Section~\ref{sec:theory-far}). The one-hidden-layer architecture also enables $O(hd)$ inference.

%We use the direct prediction form in \eqref{eq:net}, rather than a skip-connection parameterisation $f_\theta(\bx)=\bx+g_\theta(\bx)$. Although the skip-connected form explicitly parameterises the displacement, a bounded $g_\theta$ leads to a globally bounded residual score $\|g_\theta(\bx)\|$. By contrast, the direct bounded-output network yields residual scores that diverge as $\|\bx\|\to\infty$. This provides an architectural motivation for the direct parameterisation, without implying that skip-connected models necessarily fail on finite test regions. In contrast, the direct Cauchy network has bounded output and therefore produces increasing residual scores for sufficiently distant inputs, as analysed in Section~\ref{sec:theory-far}. The single-hidden-layer architecture also permits efficient inference with $O(hd)$ computational cost and provides a compact representation of the neighbourhood-mean mapping.

\subsection{Anomaly Score and Inference}
\label{sec:method-score}

The anomaly score of a query $\bx$ is the length of its predicted displacement,
\begin{equation}
s(\bx)=\left\|\bx-f_\theta(\bx)\right\|.
\label{eq:score}
\end{equation}
For a query in a well-covered normal region, a successfully fitted network is expected to predict a nearby neighbourhood mean, yielding a small residual score. Near the fringe of a normal region, larger displacements may arise from asymmetric neighbourhoods. Such behaviour is not guaranteed for every density or unseen query. For queries sufficiently far from the bounded range of the network output, the residual score necessarily grows, as established in Proposition~\ref{prop:far}. Fig.~\ref{fig:toy}(c) shows the result on the two-cluster example: the low-score region follows the clusters closely, unlike the self-reconstruction network in Fig.~\ref{fig:toy}(b), and the anomaly score increases steadily away from the data. The nonparametric mean shift in Fig.~\ref{fig:toy}(a) also follows the clusters but requires the training data and a neighbour search for every query.

Evaluating (\ref{eq:score}) needs only the parameters $\theta$ and the standardisation statistics. Inference is a batch of two matrix products and an elementwise rational function, so it vectorises well on CPUs and GPUs and is independent of the size of the training set.

\subsection{Training Procedure}
\label{sec:method-training}

\begin{algorithm}[t]
\caption{Training and scoring with \ours}
\label{alg:cdn}
\begin{algorithmic}[1]
\REQUIRE training set $\cX=\{\bx_i\}_{i=1}^n$, neighbourhood size $k$, hidden width $h$, steps $T$, batch size $b$
\STATE Standardise $\cX$ with its mean and standard deviation
\FOR{$i=1$ \TO $n$}
  \STATE $\bt_i\gets\frac{1}{k}\sum_{j\in\cN_k(\bx_i)}\bx_j$ \COMMENT{excluding $\bx_i$}
\ENDFOR
\STATE Initialise $\bW_1,\bb_1,\bW_2,\bb_2$; set $\alpha_j\gets 0.1$, $\beta_j\gets 0.2$, $\gamma_j\gets 0.6$
\FOR{$\tau=1$ \TO $T$}
  \STATE Draw a mini-batch $I$ of $\min(b,n)$ indices uniformly with replacement
  \STATE $\ell\gets\frac{1}{|I|d}\sum_{i\in I}\|f_\theta(\bx_i)-\bt_i\|^2$
  \STATE Update $\theta$ with one Adam step on $\ell$
\ENDFOR
\STATE Discard $\cX$ and the targets; keep $\theta$ and the standardisation statistics
\ENSURE scoring function $s(\bx)=\|\bx-f_\theta(\bx)\|$ for a standardised query $\bx$
\end{algorithmic}
\end{algorithm}

Algorithm~\ref{alg:cdn} summarises the procedure. Features are standardised to zero mean and unit variance with statistics of the training set, and the same transformation is applied to queries. The targets are computed with an exact $k$-nearest-neighbour search on the standardised data. The network is trained with Adam, learning rate $10^{-3}$ and weight decay $10^{-5}$ on all parameters, for a fixed budget of $T=2000$ steps with mini-batches of $\min(32,n)$ points drawn uniformly with replacement. Once the neighbour-mean targets have been computed, the neural-network optimisation cost is independent of the training-set size for fixed training steps and mini-batch size. However, the end-to-end training cost still depends on the dataset size through neighbour-search target construction. The linear layers use the default initialisation of PyTorch, and the activation parameters are initialised to $\alpha_j=0.1$, $\beta_j=0.2$ and $\gamma_j=0.6$ for every unit. We use $k=5$ and $h=128$ on all datasets, without per-dataset tuning; Section~\ref{sec:sens} shows that the results are stable over a range of these settings.

\subsection{Complexity}
\label{sec:method-complexity}

\begin{table}[t]
\centering
\caption{Cost of scoring one query and memory kept after training, assuming brute-force neighbour search. $n$: training size; $d$: dimensionality; $h$: hidden width of \ours; $I$: number of shift iterations of MOD; $|\theta|$: number of network parameters.}
\label{tab:complexity}
\small
\begin{tabular}{lcc}
\toprule
Method & Time per query & Memory at inference\\
\midrule
KNN \cite{ramaswamy2000efficient} & $O(nd)$ & $O(nd)$\\
LOF \cite{breunig2000lof} & $O(nd)$ & $O(nd)$\\
MOD \cite{yang2021mean} & $O(Ind)$ & $O(nd)$\\
LUNAR \cite{goodge2022lunar} & $O(nd+|\theta|)$ & $O(nd+|\theta|)$\\
AE \cite{aggarwal2016} & $O(|\theta|)$ & $O(|\theta|)$\\
\ours & $O(hd)$ & $O(hd)$\\
\bottomrule
\end{tabular}
\end{table}

Computing the targets requires one $k$-nearest-neighbour search over $\cX$, which costs $O(n^2d)$ with brute force and can be reduced with tree-based or approximate indices. After the targets have been computed, network optimisation costs $O(Tbhd)$, independent of the training-set size $n$ for fixed $T$ and $b$. The model has $2hd+4h+d$ parameters: $2hd+h+d$ in the linear layers and $3h$ in the activations. For $d=10$ and $h=128$ this is 3,082 parameters, whereas a neighbour-based detector trained on $n=10{,}000$ points keeps $10^5$ coordinates. Table~\ref{tab:complexity} compares the cost of scoring one query. \ours\ and AE are the only methods in the table whose inference cost does not grow with $n$, and \ours\ does so with a single hidden layer and a fixed width.

% =================================================================
\section{Theoretical Analysis}
\label{sec:theory}

This section explains the statistical meaning of the neighbourhood-mean target and the representational role of Cauchy activations in \ours. We first relate the empirical KNN displacement to graph geometry and the population density score. Student-$t$ distributions are a standard heavy-tailed model for data with outlying observations \cite{peel2000robust}, accommodating observations concentrated around characteristic regions with potentially anisotropic variation. Motivated by this connection, we consider a tractable class of Student-$t$-type expert densities and show that their score fields admit exact representations by Cauchy units. We then establish single-layer approximation guarantees and error bounds for the neighbourhood-mean mapping. Finally, we analyse anomaly-score approximation and separation, and the network's behaviour far from the training data.

\subsection{Neighbourhood Mean Shift and Density Geometry}
\label{sec:theory-target}

Let $X=(\bx_1,\ldots,\bx_n)^\top\in\R^{n\times d}$ be the matrix of $n$ normal training samples, each in $\R^d$, and let $\bt_i$ be the mean of the $k$ nearest neighbours of $\bx_i$ (excluding itself). Write $\bm_i=\bt_i-\bx_i$ for the empirical mean-shift displacement. Define $A\in\{0,1\}^{n\times n}$ as the adjacency matrix of the directed $k$-nearest-neighbour graph, $P=A/k$ as its row-stochastic transition matrix, and $L=I_n-P$ as the random-walk Laplacian, where $I_n$ is the $n\times n$ identity matrix. Then
\begin{equation}
\bx_i-\bt_i=(LX)_i .
\label{eq:laplacian}
\end{equation}
A network that fits its targets exactly therefore scores each training sample by the norm of the graph-Laplacian residual of the coordinate functions. This relates the supervision in \ours\ to local data geometry, although the magnitude of a graph-Laplacian residual does not universally distinguish high- and low-density regions.

Let $p$ denote the normal-data density, $B(\bx,r)=\{\by\in\R^d:\|\by-\bx\|<r\}$ the ball of radius $r>0$, and $T_r$ the corresponding population neighbourhood-mean map:
\begin{equation}
T_r(\bx)=\frac{\int_{B(\bx,r)}\by\,p(\by)\,d\by}{\int_{B(\bx,r)}p(\by)\,d\by},
\label{eq:population-mean}
\end{equation}
whenever the denominator is positive. Its displacement is $\bm_r(\bx)=T_r(\bx)-\bx$.

The following classical mean-shift relation \cite{fukunaga1975estimation} concerns the population mean with fixed radius $r$.

\begin{proposition}[Mean shift and density gradient]
\label{prop:meanshift}
Let $p$ be twice continuously differentiable in a neighbourhood of $\bx$ with $p(\bx)>0$. Then
\begin{equation}
T_r(\bx)-\bx=\frac{r^2}{d+2}\nabla\log p(\bx)+o(r^2),\qquad r\downarrow0.
\label{eq:meanshift-expansion}
\end{equation}
\end{proposition}
\begin{IEEEproof}
Set $\by=\bx+\bu$, where $\bu\in B(0,r)$. Taylor expansion gives $p(\bx+\bu)=p(\bx)+\nabla p(\bx)^\top\bu+O(\|\bu\|^2)$. Let $V_d$ be the volume of the unit ball in $\R^d$, and let $I_d$ be the $d\times d$ identity matrix. By symmetry,
\begin{equation*}
\int_{B(0,r)}\bu\,d\bu=0,\qquad
\int_{B(0,r)}\bu\bu^\top\,d\bu=\frac{V_dr^{d+2}}{d+2}I_d.
\end{equation*}
Consequently,
\begin{align*}
\int_{B(0,r)}\bu\,p(\bx+\bu)\,d\bu
&=\frac{V_dr^{d+2}}{d+2}\nabla p(\bx)+o(r^{d+2}),\\
\int_{B(0,r)}p(\bx+\bu)\,d\bu
&=V_dr^dp(\bx)+O(r^{d+2}).
\end{align*}
Dividing the two expressions and using $\nabla p/p=\nabla\log p$ proves the claim.
\end{IEEEproof}

Proposition~\ref{prop:meanshift} shows that the population displacement is locally proportional to the score field $\nabla\log p$. For a query $\bx$, let $r_k(\bx)$ be the distance to its $k$th nearest normal sample and $\widehat{\bm}_k(\bx)$ the displacement of the associated empirical neighbour mean. Formally replacing the fixed radius by $r_k(\bx)$ suggests $\|\widehat{\bm}_k(\bx)\|\approx r_k(\bx)^2\|\nabla\log p(\bx)\|/(d+2)$. The anomaly score of \ours\ thus combines the neighbourhood radius, which underlies KNN-based anomaly detection \cite{ramaswamy2000efficient}, and the magnitude of the density score. Neighbourhood-mean regression can therefore be interpreted as score-function estimation with data-adaptive bandwidth, conceptually related to denoising autoencoders and score matching \cite{vincent2011connection}. This interpretation is heuristic for finite-sample KNN targets, whose sampling errors are not covered by Proposition~\ref{prop:meanshift}.

\begin{remark}[Stationary points of the density]
\label{rem:critical}
The leading mean-shift term vanishes at any stationary point of $p$, not only at normal density modes. In particular, low-density gaps between symmetric clusters may have $\nabla\log p=0$. Thus a displacement-based population score need not assign a large value to every low-density point. Fig.~\ref{fig:toy} illustrates such a gap; the global far-field guarantee in Proposition~\ref{prop:far} concerns sufficiently distant queries and does not eliminate this local limitation.
\end{remark}

\subsection{Cauchy Representation of Structured Normal Densities}
\label{sec:theory-class}

In practical one-class anomaly detection, normal observations often occupy restricted characteristic regions, while their distributional parameters, directional scales, and tail behaviour are unknown. As a tractable modelling assumption for theoretical analysis, we consider a normal density constructed from Student-$t$-type factors \cite{welling2003learning}. Such factors provide heavy-tailed, direction-dependent profiles; the product family is not asserted to describe most tabular benchmarks or arbitrary skewed or multimodal densities. Crucially, \ours\ does not fit this parametric density, nor does it require its parameters to be known: the assumption identifies a structured family of score fields against which the Cauchy architecture can be analysed.

\begin{proposition}[Exact representation of Student-$t$ expert score fields]
\label{prop:pot}
Suppose the \emph{normal} density is
\begin{equation}
p(\bx)=\frac{1}{Z}\prod_{j=1}^{M}\left(1+\frac{(\mathbf w_j^\top\bx+c_j)^2}{\delta_j^2}\right)^{-\nu_j},
\label{eq:student-density}
\end{equation}
where $M\in\mathbb N$ is the number of expert factors, $\mathbf w_j\in\R^d$ are projection vectors, $c_j\in\R$ are offsets, $\nu_j>0$ are shape parameters, $\delta_j^2>\epsilon$ are squared scales, and $Z\in(0,\infty)$ is the normalising constant. Normalisability ($Z<\infty$) is assumed, not implied by $\nu_j>0$. Then $M$ Cauchy units in~\eqref{eq:net} represent $\nabla\log p$ exactly on $\R^d$.
\end{proposition}
\begin{IEEEproof}
Writing $z_j=\mathbf w_j^\top\bx+c_j$, direct differentiation gives
\begin{equation}
\nabla\log p(\bx)=-\sum_{j=1}^{M}\frac{2\nu_jz_j}{z_j^2+\delta_j^2}\,\mathbf w_j.
\label{eq:student-score}
\end{equation}
For unit $j$, choose $\alpha_j=-2\nu_j$, $\beta_j=0$, and $\gamma_j=\sqrt{\delta_j^2-\epsilon}$. Set the rows of $\bW_1$ to $\mathbf w_j^\top$, the entries of $\bb_1$ to $c_j$, $\bW_2=\bW_1^\top$, and $\bb_2=\mathbf{0}$. The resulting network output is exactly~\eqref{eq:student-score}.
\end{IEEEproof}

For example, the scalar profile $z/(z^2+\delta^2)$ is represented exactly by one Cauchy unit. On an interval containing its turning points, it cannot be represented exactly by a single affine transformation of a $\tanh$ or ReLU unit, since such
a scalar unit is monotone. This establishes a unit-level exact representation advantage rather than a general lower bound on the error or width of larger networks using other activation functions. Moreover, Proposition~\ref{prop:pot} concerns the score field, whereas \ours\ predicts the complete neighbourhood mean $T_r(\bx)$. We next bridge that distinction without modifying the direct network architecture.

\subsection{One-Layer Approximation of Neighbourhood Means}
\label{sec:theory-approx}

For sufficiently small $r>0$, Proposition~\ref{prop:meanshift} suggests the decomposition $T_r(\bx)\approx\bx+r^2\nabla\log p(\bx)/(d+2)$. The score component can be represented by Proposition~\ref{prop:pot}, while the identity component must also be realised by Cauchy units because~\eqref{eq:net} contains no explicit skip connection.

\begin{lemma}[Identity approximation]
\label{lem:id}
Let $\Omega\subset[-R,R]^d$ be a bounded input region with $R>0$. For any auxiliary Cauchy scale $\Gamma>\sqrt{\epsilon}$, there is a single-hidden-layer Cauchy network $I_\Gamma$ of width $d$ such that
\begin{equation}
\sup_{\bx\in\Omega}\|I_\Gamma(\bx)-\bx\|\le\frac{\sqrt d\,R^3}{\Gamma^2}.
\label{eq:idbound}
\end{equation}
In particular, this error can be made arbitrarily small by increasing $\Gamma$ while keeping the width equal to $d$.
\end{lemma}
\begin{IEEEproof}
The scalar function $\psi_\Gamma(z)=\Gamma^2z/(z^2+\Gamma^2)$ is a Cauchy unit with $\alpha=\Gamma^2$, $\beta=0$, and $\gamma=\sqrt{\Gamma^2-\epsilon}$. For $|z|\le R$,
\begin{equation*}
|\psi_\Gamma(z)-z|=\frac{|z|^3}{z^2+\Gamma^2}\le\frac{R^3}{\Gamma^2}.
\end{equation*}
Using one such unit per coordinate gives $I_\Gamma(\bx)=(\psi_\Gamma(x_1),\ldots,\psi_\Gamma(x_d))^\top$. Summing the squared coordinate errors proves~\eqref{eq:idbound}.
\end{IEEEproof}

\begin{proposition}[Constructive one-layer neighbourhood-mean error bound]
\label{prop:main}
Let $\Omega\subset[-R,R]^d$ be compact. Suppose $p$ satisfies~\eqref{eq:student-density} and has bounded derivatives up to order four on the closed $r_0$-neighbourhood of $\Omega$, where $r_0>0$, and is bounded away from zero there. Then there exists a finite constant $C_p>0$ (depending on $p$, $\Omega$, $r_0$, and $d$) such that for every $0<r\le r_0$ and $\Gamma>\sqrt\epsilon$, a single-hidden-layer Cauchy network $f_{r,\Gamma}$ with at most $M+d$ hidden units satisfies
\begin{equation}
\sup_{\bx\in\Omega}\|f_{r,\Gamma}(\bx)-T_r(\bx)\|\le\frac{\sqrt d\,R^3}{\Gamma^2}+C_pr^4.
\label{eq:mainbound}
\end{equation}
\end{proposition}
\begin{IEEEproof}
Put $\by=\bx+\bu$ and expand $p(\bx+\bu)$ to third order. In the numerator of $T_r(\bx)-\bx$, multiplication by $\bu$ makes the first-order Taylor term the leading nonzero contribution; the third-order term is of order $r^{d+4}$. The zeroth- and second-order terms vanish by symmetry. Uniformly for $\bx\in\Omega$,
\begin{align*}
\int_{B(0,r)}\bu p(\bx+\bu)\,d\bu
&=\frac{V_dr^{d+2}}{d+2}\nabla p(\bx)+O(r^{d+4}),\\
\int_{B(0,r)}p(\bx+\bu)\,d\bu
&=V_dr^dp(\bx)+O(r^{d+2}).
\end{align*}
Since $p$ has a uniform positive lower bound, their ratio yields
\begin{equation}
T_r(\bx)=\bx+\frac{r^2}{d+2}\nabla\log p(\bx)+\mathbf R_r(\bx),\quad
\sup_{\Omega}\|\mathbf R_r\|\le C_pr^4.
\label{eq:expand}
\end{equation}
By Proposition~\ref{prop:pot}, $g_r(\bx)=r^2\nabla\log p(\bx)/(d+2)$ is represented exactly by $M$ Cauchy units. By Lemma~\ref{lem:id}, $I_\Gamma$ uses $d$ units to approximate the identity. Concatenating the two groups of hidden units gives a direct one-hidden-layer network $f_{r,\Gamma}=I_\Gamma+g_r$ of width at most $M+d$. The triangle inequality,~\eqref{eq:idbound}, and~\eqref{eq:expand} imply~\eqref{eq:mainbound}.
\end{IEEEproof}

This proposition exploits the Student-$t$ expert structure: the score component is represented exactly, while the error arises from identity approximation and Taylor truncation. For fixed $r>0$, the upper bound need not vanish as $\Gamma\to\infty$, although this does not imply a positive lower bound on the optimal approximation error. The result is a constructive existence guarantee for a network of width at most $M+d$. It does not establish the approximation error of the fixed-width $h=128$ architecture used in experiments, nor the error of parameters obtained by gradient-based training. In particular, the theorem concerns the population fixed-radius map $T_r$ under the Student-$t$ expert assumption, not the finite-sample KNN regression targets used by CDN.

At fixed $r>0$, arbitrary accuracy follows with sufficient width. If $p$ is continuous on the closed $r$-neighbourhood of compact $\Omega$ and all radius-$r$ balls have positive probability mass, then $T_r$ is continuous. Since Cauchy activations are continuous and non-polynomial, universal approximation \cite{leshno1993multilayer} ensures that, for every $\eta>0$, some finite-width network $f_h$ satisfies
\begin{equation}
\sup_{\bx\in\Omega}\|f_h(\bx)-T_r(\bx)\|<\eta.
\label{eq:universal-mean}
\end{equation}
This qualitative result supplies neither a width-error rate nor an optimisation guarantee.

%For a fixed neighbourhood radius $r>0$, the $O(r^4)$ term in \eqref{eq:mainbound} can be avoided by allowing the network width to increase. Indeed, if $p$ is continuous on the closed $r$-neighbourhood of the compact set $\Omega$ and every radius-$r$ neighbourhood has positive probability mass, then $T_r$ is continuous. Since the Cauchy activation is continuous and non-polynomial, the universal approximation theorem for such ridge networks \cite{leshno1993multilayer} implies that, for any $\eta>0$, there exists a finite-width Cauchy network $f_h$ satisfying
%\begin{equation}
%\sup_{\bx\in\Omega}
%\|f_h(\bx)-T_r(\bx)\|<\eta.
%\label{eq:universal-mean}
%\end{equation}
%Thus, for a fixed radius and a compact domain, a Cauchy network of sufficiently large width can approximate the continuous population neighbourhood-mean mapping arbitrarily accurately. This result is qualitative and provides neither an explicit width--error rate nor a guarantee for the optimisation procedure used in experiments. This existence result does not provide an explicit convergence rate in network width or guarantee successful optimisation.

\subsection{Anomaly-Score Approximation and Separation}
\label{sec:theory-score}

For a compact region $\Omega\subset\R^d$ and an approximate neighbourhood map $f:\Omega\to\R^d$, define the population score $s_r(\bx)=\|\bx-T_r(\bx)\|$ and its approximation $s_f(\bx)=\|\bx-f(\bx)\|$.
\begin{proposition}[Score approximation and conditional separation]
\label{prop:score}
Let $\eta\ge0$. If $\sup_{\bx\in\Omega}\|f(\bx)-T_r(\bx)\|\le\eta$, then
\begin{equation}
\sup_{\bx\in\Omega}|s_f(\bx)-s_r(\bx)|\le\eta.
\label{eq:scorebound}
\end{equation}
Furthermore, if the normal region $\Omega_N$ and anomalous region $\Omega_A$ are subsets of $\Omega$ and satisfy
\begin{equation}
\sup_{\bx\in\Omega_N}s_r(\bx)\le a<b\le\inf_{\bx\in\Omega_A}s_r(\bx)
\end{equation}
for constants $a,b\in\R$ and $2\eta<b-a$, then
\begin{equation}
\sup_{\bx\in\Omega_N}s_f(\bx)<\inf_{\bx\in\Omega_A}s_f(\bx).
\label{eq:separation}
\end{equation}
\end{proposition}
\begin{IEEEproof}
The reverse triangle inequality yields
\begin{equation*}
|s_f(\bx)-s_r(\bx)|\le\|f(\bx)-T_r(\bx)\|\le\eta.
\end{equation*}
Thus $\sup_{\Omega_N}s_f\le a+\eta$ and $\inf_{\Omega_A}s_f\ge b-\eta$. The condition $2\eta<b-a$ implies strict separation.
\end{IEEEproof}

For the constructed network in Proposition~\ref{prop:main}, one can take $\eta=\sqrt d\,R^3/\Gamma^2+C_pr^4$. By the universal approximation property, a sufficiently wide Cauchy network can make $\eta$ arbitrarily small at any fixed $r>0$. Both statements concern approximation \emph{of a population score}; the positive margin $b-a$ is an additional hypothesis and is not implied by the Student-$t$ assumption alone. In particular, the stationary-point limitation of Remark~\ref{rem:critical} remains.

\subsection{Behaviour Far from the Data}
\label{sec:theory-far}

\begin{proposition}[Anomaly score equals distance to a learned centre up to a bounded term]
\label{prop:far}
Let $f_\theta$ be the direct network~\eqref{eq:net} with $h$ Cauchy units~\eqref{eq:cauchy}, and set $s_\theta(\bx)=\|\bx-f_\theta(\bx)\|$ and $\widetilde\gamma_j=\sqrt{\gamma_j^2+\epsilon}>0$. Here $\|\bW_2\|_2$ denotes the spectral norm. Define
\begin{equation}
B=\|\bW_2\|_2\sqrt h\,\max_j\left(\frac{|\alpha_j|}{2\widetilde\gamma_j}+\frac{|\beta_j|}{\widetilde\gamma_j^2}\right).
\label{eq:far-B}
\end{equation}
Then, for every $\bx\in\R^d$,
\begin{equation}
\Big|s_\theta(\bx)-\|\bx-\bb_2\|\Big|\le B.
\label{eq:far-bound}
\end{equation}
Consequently, $s_\theta(\bx)\to\infty$ as $\|\bx\|\to\infty$, and for every fixed $\bx_0$ and unit vector $\bu$,
\begin{equation}
\lim_{t\to\infty}\frac{s_\theta(\bx_0+t\bu)}{t}=1.
\label{eq:far-ray}
\end{equation}
\end{proposition}
\begin{IEEEproof}
For any $z\in\R$, $|z|/(z^2+\widetilde\gamma_j^2)\le1/(2\widetilde\gamma_j)$ and $1/(z^2+\widetilde\gamma_j^2)\le1/\widetilde\gamma_j^2$. Thus $|\phi_j(z)|\le|\alpha_j|/(2\widetilde\gamma_j)+|\beta_j|/\widetilde\gamma_j^2$, so $\|f_\theta(\bx)-\bb_2\|\le B$. The reverse triangle inequality gives~\eqref{eq:far-bound}. Dividing by $t$ along a ray and using $\|\bx_0+t\bu-\bb_2\|/t\to1$ proves~\eqref{eq:far-ray}.
\end{IEEEproof}

Proposition~\ref{prop:far} holds for any finite trained parameters without assumptions on the data distribution. It shows that the network output remains in a bounded ball centred at the learned output bias $\bb_2$, and that the residual score grows asymptotically along every fixed direction. In particular, \ours\ cannot globally reproduce the unbounded identity map. Bounded activations such as $\tanh$ share this far-field guarantee; the specific role of Cauchy units is their simultaneous structural match to the Student-$t$ expert score field in Proposition~\ref{prop:pot}.

\begin{remark}[Contrast with ReLU and skip-connected networks]\label{remark:2}
A finite ReLU network is piecewise affine and can reproduce the identity map on $\R^d$ using $2d$ ReLU units, since $x=\max(x,0)-\max(-x,0)$ coordinatewise. Consequently, no analogous architecture-wide bounded-output guarantee in Proposition~\ref{prop:far} holds for unrestricted ReLU networks. In contrast, for a skip-connected architecture $f_\theta(\bx)=\bx+g_\theta(\bx)$ with bounded $g_\theta$, the anomaly score is $\|g_\theta(\bx)\|$ and remains bounded for arbitrarily distant inputs. These effects motivate the direct bounded-output parameterisation used by \ours.
\end{remark}

\begin{figure}[!t]
\centering
\includegraphics[width=\columnwidth]{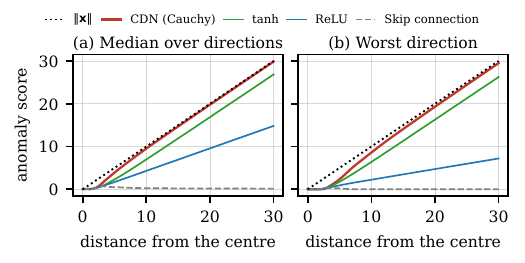}
\caption{Anomaly score far from the data on the two-cluster example of Fig.~\ref{fig:toy}. Queries lie on 360 rays from the centre of the standardised data; the horizontal axis is the distance along the ray. (a) Median score over the rays. (b) Minimum score over the rays. All networks use the configuration of Section~\ref{sec:method-training}; the dotted line is the distance itself.}
\label{fig:far}
\end{figure}

Fig.~\ref{fig:far} illustrates the far-field behaviour established in Proposition~\ref{prop:far} using the two-cluster example of Fig.~\ref{fig:toy}. Consistent with the theoretical result, the anomaly score of \ours\ exhibits approximately linear growth at large distances, with an asymptotic slope of one along every fixed direction. Over the evaluated queries, the deviation from $\|\bx-\bb_2\|$ does not exceed 2.8, well within the theoretical bound $B=29.2$ computed from the trained parameters.

The $\tanh$ network exhibits similar far-field growth because its activation is also bounded, although with a larger offset in this experiment. In contrast, the ReLU network shows slower score growth over the plotted range, with approximate slopes of 0.5 and 0.24 for the median and minimum score curves, respectively. The skip-connected network produces nearly constant scores over the same range. These observations illustrate the effect of network parameterisation on far-field scoring behaviour, without implying monotonic score growth at every finite distance.

% =================================================================
\section{Experiments}
\label{sec:exp}

\subsection{Experimental Setup}

\textbf{Datasets.} We use 46 tabular datasets from ADBench \cite{han2022adbench}. Let $N$ denote the total number of samples in each dataset before splitting, including both normal and anomalous observations, while $n$ denotes the number of normal training samples as defined in the problem formulation. The datasets range from 80 to 619{,}326 samples and from 3 to 1{,}555 features, with anomaly ratios between 0.03\% and 39.9\%; Table~\ref{tab:roc} reports $N$ and $d$ for each dataset. Following the semi-supervised protocol, the first half of the normal samples of each dataset forms the training set, and the test set contains the remaining normal samples and all anomalies. Features are standardised with statistics of the training set for all methods.

\textbf{Baselines.} We compare with 28 baselines. Classical methods: KNN \cite{ramaswamy2000efficient}, LOF \cite{breunig2000lof}, PCA \cite{shyu2003novel}, GMM \cite{aggarwal2016}, iForest \cite{liu2008isolation}, iNNE \cite{bandaragoda2018isolation}, ADERH \cite{duranianomaly}, Sampling \cite{sugiyama2013rapid}, HBOS \cite{goldstein2012histogram}, ECOD \cite{li2022ecod}, COPOD \cite{li2020copod}, QMCD \cite{fang2001wrap} and the mean-shift outlier detector MOD \cite{yang2021mean}, which is the nonparametric counterpart of \ours; we use the same $k=5$ as \ours\ and the original three shift iterations, and adapt it to the semi-supervised setting by shifting each query towards the mean of its $k$ nearest neighbours in the training set, which is kept fixed. Deep methods: AE \cite{aggarwal2016}, VAE \cite{kingma2013auto}, DeepSVDD \cite{ruff2018deep}, DAGMM \cite{zong2018deep}, ALAD \cite{zenati2018adversarially}, GANomaly \cite{akcay2018ganomaly}, SO-GAAL \cite{liu2019generative}, LUNAR \cite{goodge2022lunar}, DIF \cite{xu2023deep}, GOAD \cite{bergmanclassification}, ICL \cite{shenkar2022anomaly}, SLAD \cite{xu2023fascinating}, DDPM and DTE-C \cite{livernochediffusion}, and TCCM \cite{li2025scalable}. Implementations are taken from PyOD \cite{zhao2019pyod} and the official repositories, with the default hyperparameters of the original papers, following ADBench \cite{han2022adbench}. \ours\ uses the single configuration of Section~\ref{sec:method} on all datasets.

\textbf{Metrics.} We report AUC-ROC and AUC-PR, the latter computed as average precision. Methods with randomness are run five times with different random seeds and averaged; deterministic methods are run once. We also report the rank of each method on each dataset. Experiments run on a machine with an AMD Ryzen 9 9950X CPU, 128\,GB of memory and an NVIDIA GeForce RTX 4090 GPU.

\subsection{Overall Performance}
\label{sec:results}

\begin{table*}[!t]
\centering
\caption{AUC-ROC ($\times 100$) on 46 datasets with their total sample size $N$ and dimensionality $d$. Best in bold, second best underlined.}
\label{tab:roc}
\setlength{\tabcolsep}{1.6pt}
\resizebox{\textwidth}{!}{%
\begin{tabular}{lrr|ccccccccccccc|ccccccccccccccc|c}
\toprule
 & &  & \multicolumn{13}{c|}{Classical} & \multicolumn{15}{c|}{Deep} & \\
Dataset & $N$ & $d$ & \rotatebox{90}{KNN} & \rotatebox{90}{LOF} & \rotatebox{90}{PCA} & \rotatebox{90}{GMM} & \rotatebox{90}{iForest} & \rotatebox{90}{iNNE} & \rotatebox{90}{ADERH} & \rotatebox{90}{Sampl.} & \rotatebox{90}{HBOS} & \rotatebox{90}{ECOD} & \rotatebox{90}{COPOD} & \rotatebox{90}{QMCD} & \rotatebox{90}{MOD} & \rotatebox{90}{AE} & \rotatebox{90}{VAE} & \rotatebox{90}{DSVDD} & \rotatebox{90}{DAGMM} & \rotatebox{90}{ALAD} & \rotatebox{90}{GANom.} & \rotatebox{90}{SO-GAAL} & \rotatebox{90}{LUNAR} & \rotatebox{90}{DIF} & \rotatebox{90}{GOAD} & \rotatebox{90}{ICL} & \rotatebox{90}{SLAD} & \rotatebox{90}{DDPM} & \rotatebox{90}{DTE-C} & \rotatebox{90}{TCCM} & \rotatebox{90}{\ours} \\
\midrule
ALOI & 49534 & 27 & 51.0 & 48.8 & 54.0 & \underline{54.6} & 51.3 & 54.1 & 53.5 & 52.8 & 52.2 & 51.7 & 49.5 & 51.0 & 50.3 & 53.2 & 53.5 & 49.3 & 51.9 & 50.7 & 52.4 & \textbf{55.4} & 50.7 & 53.8 & 47.3 & 47.5 & 50.9 & 50.6 & 50.6 & 50.5 & 52.9 \\
annthyroid & 7200 & 6 & 92.8 & 88.6 & 85.2 & 84.9 & 91.0 & 92.8 & 89.1 & 88.5 & 66.0 & 78.4 & 76.8 & 73.9 & 91.8 & 82.0 & 89.2 & 67.4 & 71.3 & 55.7 & 60.7 & 76.7 & 87.5 & 74.6 & 81.2 & 78.4 & 90.4 & 75.6 & \textbf{97.5} & \underline{95.8} & 91.7 \\
backdoor & 95329 & 196 & 93.6 & 93.8 & 50.0 & 91.9 & 74.2 & 72.9 & 77.8 & 74.8 & 75.5 & 83.9 & 79.5 & 58.0 & \underline{94.4} & 92.8 & 93.8 & 93.6 & 67.1 & 48.2 & 82.8 & 87.9 & \underline{94.4} & 91.6 & 65.8 & 92.6 & 50.0 & 64.6 & 90.2 & \textbf{95.3} & 93.7 \\
breastw & 683 & 9 & 99.2 & 97.2 & 99.4 & 98.7 & \textbf{99.8} & 98.0 & 99.2 & 98.9 & \underline{99.6} & 99.5 & \textbf{99.8} & 44.4 & 98.9 & 98.9 & 99.3 & 98.1 & 95.5 & 62.1 & 95.2 & 10.1 & 99.2 & 77.6 & 98.5 & 97.5 & 99.5 & 94.5 & 91.0 & 98.1 & 99.2 \\
campaign & 41188 & 62 & 78.5 & 70.6 & 50.0 & 79.9 & 74.2 & 77.3 & 71.4 & 75.2 & 77.1 & 76.9 & 78.1 & 79.6 & 77.3 & 81.0 & 76.0 & 73.3 & 61.0 & 53.8 & 69.0 & 65.8 & 71.9 & 67.4 & 41.2 & \underline{81.2} & 76.2 & 71.2 & 79.1 & 76.1 & \textbf{83.2} \\
cardio & 1831 & 21 & 92.0 & 92.2 & \textbf{96.6} & 91.2 & 92.7 & 93.9 & 93.9 & 93.8 & 79.6 & 95.0 & 93.2 & 49.8 & 92.4 & 90.7 & 95.0 & 86.5 & 73.4 & 55.3 & 87.5 & 72.8 & 93.5 & 95.2 & \underline{96.0} & 82.3 & 80.9 & 76.0 & 88.3 & 93.9 & 94.3 \\
Cardiotocography & 2114 & 21 & 62.1 & 64.5 & 50.0 & 64.4 & \underline{74.7} & 70.4 & 57.3 & 66.3 & 61.2 & \textbf{79.3} & 66.4 & 45.0 & 61.4 & 62.8 & 73.6 & 55.4 & 58.6 & 52.7 & 62.9 & 66.8 & 62.1 & 62.7 & 74.4 & 57.2 & 47.5 & 52.8 & 61.2 & 63.7 & 66.2 \\
celeba & 202599 & 39 & 68.4 & 45.3 & 79.7 & 80.6 & 71.6 & 78.5 & \underline{80.8} & 79.3 & 76.0 & 75.7 & 75.0 & 50.0 & 64.6 & 71.6 & 71.9 & 62.7 & 66.7 & 52.8 & 32.6 & 53.3 & 62.9 & 64.7 & 29.8 & 72.0 & 65.0 & 67.1 & \textbf{82.8} & 75.5 & 73.3 \\
census & 299285 & 500 & \underline{72.0} & 60.1 & 50.0 & 70.7 & 64.0 & 71.3 & 70.0 & 71.2 & 62.4 & 66.0 & 67.5 & 68.1 & \textbf{72.2} & \underline{72.0} & 71.8 & 58.2 & 51.9 & 40.4 & 71.3 & 60.9 & 67.3 & 57.8 & 46.1 & 70.7 & 58.1 & 63.7 & 69.0 & 68.2 & 71.3 \\
cover & 286048 & 10 & 91.3 & 95.7 & 87.7 & 84.8 & 74.4 & 91.3 & 86.3 & 83.4 & 72.5 & 89.3 & 86.2 & 74.9 & 93.0 & 94.7 & 87.4 & 56.5 & 67.7 & 54.2 & 78.5 & 54.1 & 91.2 & \underline{97.5} & 19.4 & 88.6 & 80.3 & 75.0 & 96.2 & \textbf{97.7} & 94.6 \\
donors & 619326 & 10 & \underline{99.9} & 98.4 & 88.7 & 92.6 & 90.4 & 91.3 & 93.3 & 89.1 & 79.0 & 88.9 & 81.5 & 72.2 & \underline{99.9} & 92.7 & 98.3 & 70.8 & 56.5 & 55.7 & 81.9 & 33.3 & \textbf{100.0} & 89.6 & 37.6 & \textbf{100.0} & 95.2 & 82.2 & 97.9 & 82.6 & 99.3 \\
fault & 1941 & 27 & 58.7 & 47.4 & 50.0 & 61.1 & 55.8 & 58.4 & \textbf{67.2} & 60.2 & 53.1 & 50.4 & 49.1 & 53.8 & 58.8 & 60.7 & 60.3 & 58.9 & 52.0 & 51.0 & 53.6 & 47.2 & 58.7 & 62.2 & 59.3 & 60.3 & \underline{62.4} & 57.3 & 58.3 & 58.7 & 60.3 \\
fraud & 284807 & 29 & 93.2 & 35.7 & 94.4 & 93.4 & 94.0 & \underline{94.7} & 94.6 & 94.3 & 94.3 & 94.4 & 94.1 & 94.6 & 92.2 & 92.6 & 93.2 & 79.2 & 85.4 & 58.9 & 88.6 & 55.0 & 89.8 & \textbf{95.1} & 49.1 & 91.8 & 93.3 & 91.1 & 88.5 & 86.9 & 92.0 \\
glass & 214 & 7 & 84.1 & 76.9 & 68.0 & 74.2 & 76.2 & 81.8 & 85.1 & 86.0 & 83.5 & 67.5 & 73.2 & 83.2 & 86.0 & 72.0 & 69.2 & 83.2 & 63.9 & 53.1 & 65.2 & 49.2 & \underline{88.9} & 83.1 & 66.1 & \textbf{90.9} & 75.6 & 63.2 & 80.3 & 85.4 & 79.8 \\
Hepatitis & 80 & 19 & 79.2 & 81.0 & \underline{81.2} & 77.8 & 72.9 & 69.6 & 75.6 & 75.7 & 74.4 & 69.9 & 73.3 & 49.8 & \textbf{82.8} & 78.4 & 79.5 & 78.7 & 53.3 & 55.6 & 64.6 & 67.6 & 75.3 & 62.3 & 80.8 & 51.0 & 75.2 & 67.1 & 74.9 & 73.8 & 79.7 \\
http & 567498 & 3 & \textbf{99.9} & 91.1 & \underline{99.8} & \underline{99.8} & 98.9 & \textbf{99.9} & \textbf{99.9} & 99.7 & 99.4 & 97.4 & 98.9 & 0.3 & \textbf{99.9} & \underline{99.8} & \textbf{99.9} & 99.5 & 99.3 & 63.1 & 85.0 & 48.7 & \textbf{99.9} & 99.3 & 99.4 & \textbf{99.9} & 99.1 & 99.1 & 99.0 & \textbf{99.9} & \textbf{99.9} \\
InternetAds & 1966 & 1555 & 68.1 & 71.7 & 65.1 & \underline{80.1} & 47.7 & 73.3 & 62.8 & 65.8 & 49.2 & 66.0 & 65.9 & 50.0 & 71.9 & 78.1 & 78.1 & 78.2 & 50.2 & 37.3 & 69.3 & 43.0 & 78.3 & 52.1 & 65.7 & 72.4 & 75.1 & 64.6 & 78.4 & \textbf{81.6} & 72.0 \\
Ionosphere & 351 & 32 & 94.9 & 97.3 & 87.4 & 97.0 & 83.3 & 96.0 & 92.8 & 92.8 & 59.3 & 75.4 & 77.9 & 44.4 & 96.2 & 93.0 & 87.7 & \underline{97.4} & 70.5 & 51.5 & 89.6 & 73.6 & 95.4 & 89.5 & 90.8 & 95.3 & 94.5 & 81.9 & 93.3 & 96.6 & \textbf{97.6} \\
landsat & 6435 & 36 & 68.2 & 66.6 & 43.9 & 45.2 & 59.7 & 62.3 & 54.6 & 55.7 & \textbf{73.2} & 42.0 & 49.3 & 63.0 & \underline{69.4} & 53.2 & 53.6 & 59.8 & 51.3 & 49.0 & 54.7 & 48.7 & 69.0 & 56.4 & 39.8 & 65.0 & 63.7 & 49.5 & 51.0 & 53.9 & 62.3 \\
letter & 1600 & 32 & 35.4 & 44.8 & 30.2 & 35.1 & 30.9 & 36.1 & 36.3 & 32.1 & 35.9 & 45.4 & 36.5 & \underline{57.6} & 36.9 & 34.4 & 31.8 & 41.9 & 47.1 & 45.8 & 36.3 & 38.7 & 37.6 & \textbf{62.8} & 30.9 & 45.1 & 36.7 & 35.7 & 37.4 & 35.8 & 35.8 \\
Lymphography & 148 & 18 & \textbf{100.0} & \textbf{100.0} & \textbf{100.0} & 94.6 & 99.7 & \textbf{100.0} & \textbf{100.0} & \textbf{100.0} & 98.1 & 99.1 & 99.3 & 2.6 & \textbf{100.0} & 94.2 & 93.9 & 99.8 & 91.5 & 49.1 & 98.4 & 2.6 & 97.2 & 90.5 & \underline{99.9} & 98.4 & \textbf{100.0} & 99.6 & \underline{99.9} & 99.8 & \textbf{100.0} \\
magic.gamma & 19020 & 10 & 83.3 & 83.4 & 70.4 & 80.1 & 77.6 & 78.6 & 68.7 & 73.8 & 74.5 & 63.6 & 68.0 & 71.4 & 83.5 & 82.1 & 78.1 & 72.2 & 62.4 & 53.5 & 60.7 & 50.7 & \underline{85.2} & 76.2 & 69.4 & 76.9 & 70.4 & 69.6 & \textbf{87.2} & 71.0 & 83.9 \\
mammography & 11183 & 6 & 87.6 & 85.5 & 89.7 & 88.1 & 88.3 & 83.9 & 86.9 & 84.4 & 85.0 & \textbf{90.7} & \underline{90.6} & 70.9 & 87.4 & 89.2 & 88.7 & 88.2 & 83.3 & 56.5 & 84.8 & 78.7 & 88.2 & 82.7 & 73.3 & 78.2 & 65.0 & 72.6 & 85.2 & 82.1 & 87.7 \\
mnist & 7603 & 100 & \underline{93.8} & 92.9 & 50.0 & 92.1 & 87.0 & 92.1 & 90.0 & 90.3 & 62.4 & 74.9 & 77.7 & 64.4 & \textbf{94.5} & 92.9 & 91.3 & 80.8 & 69.2 & 52.2 & 76.1 & 65.8 & \underline{93.8} & 87.9 & 91.0 & 90.4 & 73.7 & 80.1 & 88.1 & 91.3 & 93.7 \\
musk & 3062 & 166 & \textbf{100.0} & \textbf{100.0} & \textbf{100.0} & \underline{99.9} & 97.0 & \textbf{100.0} & \textbf{100.0} & \textbf{100.0} & \textbf{100.0} & \underline{99.9} & 99.7 & 18.8 & \textbf{100.0} & \textbf{100.0} & \textbf{100.0} & 99.5 & 91.9 & 59.2 & \textbf{100.0} & 97.5 & \textbf{100.0} & \textbf{100.0} & \textbf{100.0} & 99.0 & \textbf{100.0} & \textbf{100.0} & \textbf{100.0} & \textbf{100.0} & \textbf{100.0} \\
optdigits & 5216 & 64 & 93.7 & 96.6 & 50.0 & 81.5 & 84.5 & 74.8 & 63.4 & 73.1 & 89.9 & 61.6 & 69.3 & 12.9 & 95.6 & 87.4 & 78.6 & 74.4 & 48.0 & 53.1 & 67.8 & 48.2 & \textbf{99.5} & 61.1 & 65.0 & \underline{96.9} & 92.6 & 62.2 & 81.8 & 82.9 & 86.6 \\
PageBlocks & 5393 & 10 & 89.7 & \underline{91.3} & 86.2 & 88.6 & 82.6 & \textbf{93.0} & 88.2 & 89.0 & 65.6 & 88.0 & 80.9 & 42.7 & 88.7 & 88.7 & 89.2 & 87.4 & 76.1 & 58.7 & 73.2 & 73.3 & 88.5 & 88.5 & 88.2 & 88.5 & 87.5 & 83.7 & 90.2 & 87.7 & 90.7 \\
pendigits & 6870 & 16 & \textbf{99.9} & 99.1 & 94.4 & 85.0 & 96.6 & 93.6 & 96.6 & 93.0 & 93.6 & 93.0 & 90.7 & 90.6 & \textbf{99.9} & 97.7 & 95.1 & 92.5 & 70.1 & 53.7 & 67.3 & 65.1 & \textbf{99.9} & 95.9 & 87.6 & 96.2 & 92.9 & 84.5 & 98.0 & 97.8 & \underline{99.5} \\
Pima & 768 & 8 & \underline{74.3} & 67.4 & 70.8 & 72.1 & 73.4 & 72.4 & 71.5 & 69.9 & 73.9 & 59.1 & 65.6 & \textbf{76.0} & 71.1 & 70.9 & 71.3 & 69.4 & 57.8 & 51.3 & 62.0 & 31.3 & 74.2 & 63.1 & 70.0 & 66.0 & 57.5 & 57.4 & 65.0 & 54.4 & 71.5 \\
satellite & 6435 & 36 & 82.2 & 80.3 & 66.6 & 78.7 & 77.2 & 78.8 & 76.3 & 79.0 & \underline{85.5} & 62.2 & 68.4 & 79.7 & 82.6 & 78.1 & 76.5 & 82.5 & 61.1 & 51.8 & 72.6 & 65.1 & 82.3 & 77.0 & 68.8 & \underline{85.5} & \textbf{86.6} & 77.8 & 78.4 & 77.1 & 78.5 \\
satimage-2 & 5803 & 36 & 99.7 & 99.4 & 98.2 & 99.7 & 99.1 & \underline{99.8} & \underline{99.8} & 99.4 & 98.0 & 97.1 & 97.9 & 99.2 & 99.6 & \textbf{99.9} & \underline{99.8} & 97.3 & 73.7 & 56.6 & 97.7 & 90.6 & 99.3 & 99.7 & 99.0 & 99.5 & \underline{99.8} & 98.4 & 99.4 & 99.1 & 99.7 \\
shuttle & 49097 & 9 & \underline{99.9} & \textbf{100.0} & 99.4 & 99.4 & 99.6 & \underline{99.9} & 99.8 & 99.6 & 98.6 & 99.3 & 99.5 & 72.2 & \underline{99.9} & 99.8 & \underline{99.9} & 99.7 & 91.7 & 65.8 & 98.0 & 11.0 & \textbf{100.0} & 99.1 & 88.8 & \textbf{100.0} & \underline{99.9} & 99.8 & 99.8 & \underline{99.9} & \underline{99.9} \\
skin & 245057 & 3 & \textbf{91.7} & 65.7 & 61.0 & 87.8 & 88.1 & 90.2 & 89.7 & 91.1 & 75.3 & 56.4 & 53.3 & 79.0 & 90.5 & 81.4 & 87.6 & 75.6 & 55.0 & 52.2 & 49.4 & 61.9 & 90.4 & 83.7 & 75.5 & 5.0 & \underline{91.3} & 82.4 & 89.9 & 60.8 & 89.4 \\
smtp & 95156 & 3 & 91.4 & 85.7 & 83.8 & 80.7 & 91.2 & 92.3 & 92.4 & 90.2 & 83.2 & 88.4 & 92.0 & 43.5 & 89.1 & 88.7 & 87.1 & \underline{94.3} & 84.8 & 60.5 & 49.2 & 56.1 & 91.6 & 83.7 & 92.0 & 78.7 & 78.6 & 86.9 & \textbf{94.9} & 85.4 & 88.1 \\
SpamBase & 4207 & 57 & 83.4 & 73.2 & 81.4 & 80.7 & \underline{84.6} & 82.1 & 81.0 & 81.6 & 77.9 & 68.8 & 72.1 & 71.2 & 83.9 & 82.3 & 81.7 & 82.9 & 66.7 & 56.2 & 82.5 & 44.4 & 81.4 & 54.6 & 81.5 & 83.4 & \textbf{85.2} & 79.0 & 82.9 & 79.9 & 84.2 \\
speech & 3686 & 400 & 36.4 & 37.5 & 36.4 & 37.8 & 38.3 & 38.8 & 43.1 & 36.8 & 36.7 & 36.0 & 37.0 & \textbf{55.4} & 39.4 & 36.3 & 36.4 & 45.6 & 47.1 & \underline{50.0} & 39.1 & 45.8 & 44.1 & 39.8 & 37.1 & 48.2 & 41.2 & 45.9 & 37.7 & 39.7 & 37.4 \\
Stamps & 340 & 9 & 92.4 & 91.6 & 91.3 & 90.9 & 91.5 & 91.9 & 91.4 & 90.0 & 91.1 & 89.7 & \textbf{93.4} & 88.5 & 92.8 & 89.2 & 91.5 & 89.5 & 57.5 & 54.4 & 78.0 & 70.6 & 91.1 & 90.3 & 80.6 & 79.5 & 72.1 & 76.4 & 89.9 & 87.3 & \underline{93.0} \\
thyroid & 3772 & 6 & 98.7 & 92.7 & 98.6 & 98.2 & \textbf{99.0} & 98.7 & 98.6 & 98.2 & 98.6 & 97.6 & 93.8 & 67.5 & 98.3 & 96.8 & 97.6 & 93.8 & 83.9 & 62.9 & 94.3 & 96.2 & 98.2 & 98.5 & 95.9 & 94.4 & 95.5 & 91.3 & \underline{98.9} & 96.1 & 98.6 \\
vertebral & 240 & 6 & 16.8 & 30.7 & 17.5 & 23.5 & 17.1 & 32.3 & 32.7 & 22.1 & 7.2 & 41.2 & 27.4 & 8.1 & 29.8 & 22.4 & 17.1 & 33.3 & 35.1 & 49.8 & 17.3 & \underline{53.3} & 21.0 & \textbf{58.3} & 34.0 & 44.7 & 47.7 & 38.4 & 45.7 & 37.7 & 33.2 \\
vowels & 1456 & 12 & 82.2 & \underline{86.3} & 52.8 & 83.2 & 62.4 & 85.9 & 84.8 & 70.6 & 53.3 & 61.5 & 52.8 & 46.1 & 81.6 & 83.0 & 75.0 & 74.5 & 40.2 & 48.8 & 50.8 & 26.0 & 82.6 & 70.4 & 65.6 & 82.9 & 80.6 & 70.3 & \textbf{87.7} & 80.2 & 81.3 \\
Waveform & 3443 & 21 & 75.2 & 76.0 & 65.0 & 58.5 & 71.7 & \underline{77.4} & 69.4 & 67.7 & 69.3 & 59.4 & 72.4 & 73.7 & 71.4 & 65.5 & 76.1 & 65.4 & 51.0 & 51.0 & \textbf{78.9} & 38.9 & 75.1 & 71.7 & 66.5 & 67.1 & 50.1 & 50.0 & 64.0 & 53.5 & 69.7 \\
WBC & 223 & 9 & 98.5 & 96.3 & 99.4 & 97.2 & \textbf{99.9} & 94.8 & 99.1 & 98.7 & \underline{99.8} & \underline{99.8} & \underline{99.8} & 37.5 & 97.4 & 98.6 & 99.2 & 95.9 & 93.0 & 59.5 & 92.1 & 19.1 & 98.5 & 76.4 & 98.6 & 91.4 & 99.4 & 90.3 & 70.0 & 97.6 & 98.6 \\
WDBC & 367 & 30 & 99.8 & \underline{99.9} & \underline{99.9} & \underline{99.9} & \underline{99.9} & 99.8 & 99.8 & 99.7 & 99.3 & 97.9 & 99.8 & 51.7 & 99.7 & 99.8 & \underline{99.9} & 99.8 & 70.6 & 55.1 & 89.7 & 13.7 & 99.8 & 78.4 & \underline{99.9} & 99.1 & 99.0 & 84.7 & 94.5 & \textbf{100.0} & 99.7 \\
Wilt & 4819 & 5 & 63.7 & 68.8 & 46.4 & \textbf{85.2} & 48.8 & 40.8 & 39.4 & 45.3 & 39.1 & 37.5 & 32.1 & 37.9 & 68.5 & 58.9 & 43.0 & 73.5 & 50.3 & 48.2 & 51.0 & 46.9 & 60.4 & 42.8 & 55.3 & 76.7 & 65.2 & 53.2 & \underline{85.0} & 46.6 & 79.7 \\
wine & 129 & 13 & 46.7 & 40.8 & 43.8 & 35.5 & 46.7 & 55.8 & 61.3 & 51.4 & 41.5 & 74.3 & \textbf{87.3} & 52.8 & 39.7 & 42.3 & 44.7 & 30.4 & 40.9 & 47.6 & 54.0 & 12.2 & 47.3 & 74.3 & 42.7 & \underline{81.9} & 45.3 & 42.0 & 39.4 & 28.0 & 43.7 \\
WPBC & 198 & 33 & 49.5 & 46.5 & 46.8 & 46.5 & 49.0 & 47.6 & 47.5 & 49.9 & 51.4 & 47.1 & 50.7 & 45.6 & 48.9 & 46.8 & 45.9 & 47.5 & \textbf{55.5} & 49.4 & 49.9 & 52.2 & 51.0 & 49.4 & 48.4 & 45.2 & \underline{52.3} & 49.5 & 52.1 & 45.7 & 48.6 \\
\midrule
Average & & & 80.7 & 77.3 & 72.0 & 78.8 & 76.7 & 79.5 & 78.5 & 77.8 & 73.3 & 74.7 & 74.9 & 57.0 & \underline{81.0} & 78.9 & 78.4 & 76.0 & 65.3 & 53.3 & 69.9 & 52.7 & 80.7 & 75.9 & 69.4 & 78.1 & 76.1 & 71.9 & 79.8 & 77.2 & \textbf{81.2} \\
Avg. rank & & & 9.28 & 12.85 & 16.76 & 13.09 & 13.88 & 9.77 & 11.78 & 12.96 & 16.45 & 17.01 & 16.48 & 20.82 & \underline{9.26} & 12.40 & 11.68 & 15.72 & 22.26 & 24.61 & 19.43 & 23.12 & 9.85 & 15.37 & 18.12 & 13.23 & 13.96 & 20.15 & 12.07 & 14.00 & \textbf{8.65} \\
\bottomrule
\end{tabular}}
\end{table*}

\begin{table*}[!t]
\centering
\caption{AUC-PR ($\times 100$) on 46 datasets. Best in bold, second best underlined.}
\label{tab:pr}
\setlength{\tabcolsep}{1.6pt}
\resizebox{\textwidth}{!}{%
\begin{tabular}{lccccccccccccc|ccccccccccccccc|c}
\toprule
 & \multicolumn{13}{c|}{Classical} & \multicolumn{15}{c|}{Deep} & \\
Dataset & \rotatebox{90}{KNN} & \rotatebox{90}{LOF} & \rotatebox{90}{PCA} & \rotatebox{90}{GMM} & \rotatebox{90}{iForest} & \rotatebox{90}{iNNE} & \rotatebox{90}{ADERH} & \rotatebox{90}{Sampl.} & \rotatebox{90}{HBOS} & \rotatebox{90}{ECOD} & \rotatebox{90}{COPOD} & \rotatebox{90}{QMCD} & \rotatebox{90}{MOD} & \rotatebox{90}{AE} & \rotatebox{90}{VAE} & \rotatebox{90}{DSVDD} & \rotatebox{90}{DAGMM} & \rotatebox{90}{ALAD} & \rotatebox{90}{GANom.} & \rotatebox{90}{SO-GAAL} & \rotatebox{90}{LUNAR} & \rotatebox{90}{DIF} & \rotatebox{90}{GOAD} & \rotatebox{90}{ICL} & \rotatebox{90}{SLAD} & \rotatebox{90}{DDPM} & \rotatebox{90}{DTE-C} & \rotatebox{90}{TCCM} & \rotatebox{90}{\ours} \\
\midrule
ALOI & 6.0 & 6.5 & 6.5 & 6.5 & 5.9 & 6.5 & 6.4 & 6.3 & 6.4 & 6.1 & 5.7 & 6.0 & 6.0 & 6.2 & 6.4 & 5.7 & 6.3 & 6.2 & \textbf{7.6} & 6.9 & 6.7 & \underline{7.2} & 5.6 & 5.5 & 5.9 & 6.0 & 5.8 & 5.9 & 6.3 \\
annthyroid & 68.1 & 53.5 & 56.4 & 56.0 & 61.2 & 69.2 & 59.5 & 62.8 & 39.0 & 40.0 & 29.6 & 35.5 & 65.2 & 58.5 & 58.5 & 37.0 & 29.7 & 17.6 & 22.7 & 32.3 & 55.7 & 46.1 & 59.8 & 41.8 & 66.9 & 47.1 & \textbf{82.9} & \underline{74.8} & 63.6 \\
backdoor & 48.0 & 37.4 & 4.8 & 85.8 & 9.0 & 15.7 & 14.4 & 12.3 & 10.5 & 15.5 & 13.1 & 5.8 & 50.5 & \textbf{86.0} & \underline{85.9} & 83.3 & 9.0 & 4.9 & 16.2 & 25.1 & 69.8 & 56.1 & 13.3 & 79.4 & 4.8 & 10.3 & 60.7 & 67.9 & 55.6 \\
breastw & 99.0 & 94.5 & 99.4 & 98.3 & \underline{99.7} & 97.8 & 98.9 & 98.6 & 99.6 & 99.6 & \textbf{99.8} & 63.7 & 98.5 & 98.4 & 99.2 & 97.5 & 95.5 & 65.2 & 94.9 & 36.9 & 99.0 & 68.3 & 98.6 & 96.3 & 99.5 & 95.2 & 87.6 & 98.1 & 99.2 \\
campaign & 49.0 & 40.2 & 20.2 & \underline{51.4} & 45.3 & 47.3 & 46.0 & 48.7 & 49.7 & 49.5 & 51.0 & 48.4 & 46.1 & 49.4 & 44.4 & 44.0 & 32.8 & 23.5 & 36.9 & 40.6 & 41.8 & 37.8 & 20.9 & 49.4 & 47.8 & 45.1 & 48.4 & 46.6 & \textbf{53.7} \\
cardio & 77.2 & 70.2 & \textbf{86.3} & 72.7 & 77.7 & 78.2 & 80.7 & 81.3 & 58.4 & 78.6 & 74.9 & 33.3 & 77.3 & 75.0 & 82.2 & 66.9 & 47.4 & 24.9 & 69.3 & 48.7 & 80.3 & 80.8 & \underline{85.0} & 57.2 & 69.0 & 57.7 & 70.5 & 82.7 & 82.6 \\
Cardiotocography & 57.4 & 57.3 & 36.1 & 55.8 & 63.5 & 61.2 & 55.5 & 61.1 & 50.7 & \textbf{69.0} & 56.1 & 36.1 & 57.8 & 56.8 & 65.2 & 51.4 & 48.4 & 39.5 & 56.0 & 55.8 & 58.6 & 57.4 & \underline{66.3} & 50.5 & 49.2 & 48.6 & 53.9 & 59.7 & 60.5 \\
celeba & 9.9 & 3.8 & \textbf{20.9} & 16.4 & 12.3 & 14.0 & \underline{18.9} & 18.8 & 16.9 & 17.0 & 16.6 & 4.4 & 7.8 & 10.0 & 11.5 & 8.6 & 9.9 & 5.1 & 3.3 & 4.7 & 6.9 & 8.9 & 3.3 & 10.3 & 7.6 & 9.8 & 15.8 & 17.0 & 11.4 \\
census & 21.1 & 14.1 & 11.7 & 20.1 & 14.6 & 20.8 & 19.9 & 20.2 & 13.9 & 15.5 & 16.2 & 20.8 & \textbf{21.7} & \underline{21.6} & 21.0 & 14.8 & 12.8 & 9.5 & 19.8 & 16.0 & 16.4 & 12.6 & 12.6 & 21.4 & 15.2 & 18.0 & 17.6 & 18.2 & 20.7 \\
cover & 17.8 & \underline{41.8} & 7.1 & 6.5 & 4.2 & 15.6 & 7.1 & 7.4 & 5.0 & 14.4 & 9.8 & 5.0 & 23.6 & 27.2 & 13.1 & 2.8 & 3.5 & 2.4 & 17.3 & 2.1 & 25.7 & 40.2 & 1.1 & 25.5 & 6.5 & 5.0 & 32.2 & \textbf{74.9} & 27.7 \\
donors & 97.2 & 77.7 & 36.3 & 43.3 & 42.7 & 40.7 & 46.8 & 38.5 & 25.7 & 41.5 & 33.6 & 16.8 & 98.0 & 47.2 & 80.1 & 30.8 & 15.6 & 14.8 & 33.0 & 9.0 & \textbf{100.0} & 34.4 & 9.5 & \underline{99.5} & 59.4 & 31.7 & 70.2 & 34.4 & 88.8 \\
fault & 62.0 & 50.4 & 51.5 & 63.1 & 59.0 & 61.4 & \textbf{66.6} & 61.6 & 53.9 & 51.7 & 53.2 & 54.8 & 62.4 & 64.0 & 63.3 & 62.9 & 56.7 & 52.7 & 57.6 & 50.6 & 62.6 & 64.4 & 62.6 & 63.0 & 64.7 & 59.2 & 63.2 & \underline{65.0} & 64.5 \\
fraud & 25.4 & 0.3 & 23.7 & 40.9 & 17.8 & 35.5 & 23.1 & 24.6 & 34.4 & 31.6 & 33.8 & 39.1 & 26.0 & 57.0 & 32.2 & 37.6 & 14.0 & 1.2 & 46.1 & 0.4 & 44.0 & 52.6 & 17.8 & 45.4 & 35.2 & \underline{73.5} & \textbf{80.0} & 70.5 & 36.4 \\
glass & 25.8 & 19.4 & 18.6 & 25.9 & 19.9 & 29.8 & 32.4 & 28.1 & 26.6 & 23.4 & 26.3 & 24.0 & 27.3 & 21.3 & 18.6 & 32.7 & 17.6 & 12.8 & 14.8 & 13.9 & \underline{39.2} & 33.2 & 19.8 & \textbf{47.8} & 25.2 & 21.1 & 26.3 & 34.0 & 23.1 \\
Hepatitis & 59.7 & \underline{61.3} & 58.3 & 47.0 & 42.4 & 42.2 & 48.0 & 52.5 & 44.5 & 40.5 & 46.6 & 35.5 & \textbf{63.0} & 48.8 & 49.7 & 59.8 & 37.9 & 35.5 & 45.6 & 52.9 & 49.8 & 38.0 & 58.8 & 30.0 & 52.7 & 48.8 & 54.0 & 44.6 & 57.8 \\
http & 85.6 & 7.7 & 75.7 & 75.2 & 46.8 & 82.0 & 79.8 & 70.7 & 56.4 & 22.2 & 39.0 & 0.7 & 85.9 & 78.0 & 88.5 & 68.7 & 42.8 & 3.4 & 22.8 & 1.1 & 85.8 & 43.8 & 54.1 & \textbf{92.4} & 44.1 & 27.0 & 41.8 & \underline{91.8} & 85.8 \\
InternetAds & 49.2 & 50.4 & 47.0 & 66.3 & 29.4 & 60.7 & 45.8 & 47.5 & 30.8 & 61.9 & 61.7 & 31.5 & 52.4 & \underline{69.2} & \underline{69.2} & 65.2 & 31.9 & 25.6 & 49.9 & 31.5 & 68.2 & 32.2 & 48.2 & 60.6 & 58.2 & 47.2 & 59.0 & \textbf{74.2} & 52.1 \\
Ionosphere & 95.9 & \textbf{98.2} & 89.5 & 97.9 & 84.4 & 97.2 & 94.4 & 93.9 & 53.2 & 78.6 & 77.9 & 46.3 & 96.8 & 94.7 & 89.7 & \underline{98.0} & 74.2 & 57.6 & 89.1 & 80.6 & 96.5 & 91.1 & 92.5 & 96.2 & 96.4 & 85.7 & 95.8 & 97.6 & \textbf{98.2} \\
landsat & 54.8 & \textbf{61.4} & 32.7 & 31.7 & 47.7 & 51.6 & 35.7 & 35.7 & \underline{60.1} & 31.1 & 33.8 & 46.4 & 58.5 & 39.3 & 40.5 & 47.5 & 38.2 & 33.2 & 36.6 & 32.4 & 56.6 & 44.7 & 30.9 & 53.0 & 44.8 & 32.8 & 36.2 & 47.9 & 57.5 \\
letter & 8.7 & 11.3 & 8.0 & 8.7 & 8.0 & 8.6 & 8.8 & 8.2 & 8.7 & 10.6 & 8.8 & \underline{19.5} & 8.9 & 8.5 & 8.2 & 10.7 & 12.6 & 11.0 & 9.3 & 10.0 & 8.9 & \textbf{24.3} & 8.1 & 12.4 & 8.9 & 8.7 & 9.0 & 8.7 & 8.8 \\
Lymphography & \textbf{100.0} & \textbf{100.0} & \textbf{100.0} & 62.2 & 96.6 & \textbf{100.0} & \textbf{100.0} & \textbf{100.0} & 76.6 & 89.7 & 91.5 & 4.7 & \textbf{100.0} & 41.4 & 40.9 & 97.7 & 70.3 & 13.7 & 84.2 & 4.7 & 79.4 & 57.4 & \underline{99.2} & 86.1 & \textbf{100.0} & 97.6 & \underline{99.2} & 97.6 & \textbf{100.0} \\
magic.gamma & 85.9 & 86.4 & 75.2 & 83.3 & 80.5 & 82.4 & 74.8 & 78.6 & 77.1 & 67.9 & 72.2 & 75.3 & 86.3 & 85.4 & 81.6 & 76.7 & 67.0 & 56.1 & 66.8 & 54.0 & \underline{87.7} & 80.7 & 76.6 & 82.2 & 76.0 & 73.8 & \textbf{89.4} & 76.8 & 86.5 \\
mammography & 41.3 & 34.1 & 42.8 & 41.0 & 37.4 & 35.4 & 38.9 & 33.1 & 21.3 & \textbf{55.2} & \underline{54.6} & 14.1 & 41.1 & 41.8 & 43.7 & 50.6 & 39.9 & 7.5 & 24.6 & 16.8 & 46.4 & 34.4 & 29.3 & 20.3 & 13.4 & 11.9 & 37.0 & 28.7 & 43.1 \\
mnist & 72.7 & 71.0 & 16.9 & 69.9 & 54.8 & 68.5 & 64.7 & 65.4 & 22.2 & 30.2 & 35.4 & 37.1 & \underline{75.7} & 70.8 & 68.0 & 51.7 & 43.4 & 20.0 & 46.0 & 37.5 & \textbf{79.3} & 61.8 & 66.3 & 69.4 & 47.6 & 52.6 & 57.7 & 68.6 & 72.9 \\
musk & \textbf{100.0} & \textbf{100.0} & \textbf{100.0} & 98.5 & 64.3 & \textbf{100.0} & 99.6 & \textbf{100.0} & \textbf{100.0} & 98.2 & 96.1 & 3.8 & \textbf{100.0} & \textbf{100.0} & \textbf{100.0} & 94.8 & 68.0 & 15.0 & \textbf{100.0} & 90.7 & \textbf{100.0} & \underline{99.7} & \textbf{100.0} & 88.6 & \textbf{100.0} & \underline{99.7} & \textbf{100.0} & \textbf{100.0} & \textbf{100.0} \\
optdigits & 29.1 & 43.6 & 5.6 & 13.4 & 17.3 & 10.2 & 6.8 & 9.3 & 42.4 & 6.7 & 8.3 & 3.0 & 36.6 & 18.0 & 12.0 & 14.5 & 5.4 & 7.0 & 8.9 & 6.0 & \textbf{86.8} & 7.5 & 7.3 & \underline{49.2} & 28.4 & 7.7 & 15.1 & 17.2 & 17.0 \\
PageBlocks & 67.6 & \underline{71.1} & 59.5 & 62.8 & 43.5 & \textbf{74.7} & 62.2 & 66.2 & 22.5 & 58.5 & 41.5 & 14.7 & 67.6 & 66.1 & 67.0 & 57.7 & 48.6 & 25.2 & 46.0 & 41.3 & 68.3 & 65.6 & 63.8 & 70.3 & 64.2 & 60.5 & 68.2 & 67.2 & 68.0 \\
pendigits & \underline{97.0} & 78.6 & 39.2 & 15.9 & 52.8 & 35.5 & 60.5 & 35.5 & 42.3 & 41.4 & 30.9 & 27.4 & 96.9 & 55.7 & 51.0 & 50.5 & 18.6 & 5.8 & 15.5 & 8.5 & \textbf{98.8} & 52.7 & 31.7 & 62.5 & 36.3 & 24.7 & 49.9 & 76.7 & 89.2 \\
Pima & 71.7 & 66.6 & 69.7 & 70.0 & 72.2 & 70.5 & 69.7 & 69.7 & \underline{74.0} & 63.4 & 68.3 & \textbf{75.5} & 71.9 & 70.1 & 69.8 & 69.7 & 58.2 & 54.4 & 64.2 & 41.6 & 71.9 & 59.8 & 70.0 & 67.8 & 60.1 & 61.3 & 64.2 & 61.5 & 72.2 \\
satellite & 86.0 & 85.8 & 77.8 & 84.6 & 82.4 & 83.2 & 81.7 & 81.6 & \underline{86.5} & 69.6 & 73.3 & 82.5 & 86.4 & 84.4 & 83.1 & 85.7 & 64.8 & 51.0 & 75.5 & 70.5 & 85.9 & 83.1 & 79.1 & \textbf{87.8} & \textbf{87.8} & 83.2 & 84.3 & 83.3 & 84.8 \\
satimage-2 & 96.7 & 88.5 & 91.9 & 85.1 & 94.5 & 96.6 & \underline{97.3} & 96.5 & 87.7 & 79.7 & 85.3 & 94.9 & 96.3 & 97.0 & \textbf{97.6} & 89.4 & 16.1 & 6.0 & 84.8 & 17.9 & 94.7 & 95.6 & 95.9 & 95.0 & 96.2 & 75.5 & 70.1 & 95.4 & 97.1 \\
shuttle & 97.9 & \textbf{99.8} & 96.3 & 96.8 & 98.6 & \underline{98.8} & 97.0 & 96.5 & 97.5 & 95.2 & 98.0 & 69.7 & 98.2 & 97.5 & 97.6 & 98.2 & 71.8 & 31.6 & 96.7 & 8.1 & \textbf{99.8} & 98.1 & 77.9 & \textbf{99.8} & 97.3 & 97.5 & 94.2 & 98.6 & 97.8 \\
skin & 70.1 & 39.1 & 39.9 & 61.7 & 62.0 & 67.0 & 65.0 & 68.5 & 53.7 & 33.8 & 32.1 & 64.6 & 68.1 & 55.4 & 65.0 & 52.9 & 37.8 & 36.5 & 34.4 & 44.7 & \underline{70.5} & 55.3 & 46.7 & 33.7 & \textbf{82.5} & 64.0 & 65.3 & 42.5 & 65.7 \\
smtp & 44.7 & 21.9 & 44.7 & 36.0 & 2.1 & 49.4 & 48.6 & 45.0 & 1.1 & 55.8 & 1.1 & 1.4 & 44.7 & 27.4 & 40.7 & 30.0 & 46.5 & 0.5 & 0.1 & 0.1 & \textbf{66.8} & \underline{61.2} & 41.4 & 14.4 & 35.7 & 25.4 & 44.0 & 43.8 & 44.5 \\
SpamBase & 83.3 & 72.7 & 81.8 & 81.2 & \textbf{87.8} & 82.3 & 82.3 & 82.1 & 78.4 & 71.3 & 73.6 & 78.1 & 83.8 & 82.5 & 82.1 & 82.5 & 73.0 & 63.9 & 83.7 & 59.4 & 81.2 & 60.7 & 82.0 & \underline{86.6} & 86.2 & 80.6 & 83.4 & 80.4 & 83.9 \\
speech & 2.8 & 3.2 & 2.8 & 3.0 & 2.9 & 2.6 & 2.9 & 2.8 & 3.2 & 2.9 & 2.8 & \textbf{3.9} & 2.7 & 2.7 & 2.8 & 3.1 & 3.3 & 3.4 & 2.7 & 3.0 & 3.4 & 3.1 & 2.7 & \underline{3.8} & 2.8 & 3.2 & 2.8 & 2.9 & 2.8 \\
Stamps & 57.4 & 54.9 & 54.1 & 52.7 & 51.5 & 57.2 & 54.2 & 54.3 & 52.1 & 53.6 & 57.1 & 51.6 & \underline{58.7} & 52.2 & 54.8 & 51.0 & 22.3 & 23.7 & 43.2 & 38.0 & 53.5 & 52.1 & 47.2 & 38.3 & 32.3 & 43.8 & 54.0 & 49.4 & \textbf{58.8} \\
thyroid & 80.9 & 60.6 & 81.4 & 80.2 & 78.6 & 80.2 & 79.5 & 79.8 & 77.0 & 64.0 & 30.2 & 37.3 & 80.0 & 69.8 & 72.9 & 77.5 & 19.7 & 10.8 & 52.4 & 53.4 & 78.2 & \underline{82.6} & 80.7 & 50.7 & 71.3 & 71.8 & \textbf{83.3} & 81.8 & 82.1 \\
vertebral & 13.7 & 16.4 & 13.8 & 14.7 & 13.7 & 17.2 & 16.2 & 14.6 & 12.8 & 19.2 & 15.3 & 12.9 & 15.9 & 14.6 & 13.8 & 17.1 & 17.7 & 24.6 & 14.1 & \textbf{28.1} & 14.4 & \underline{25.7} & 16.7 & 21.8 & 21.1 & 18.3 & 20.3 & 17.7 & 16.6 \\
vowels & 30.2 & 33.1 & 10.6 & \underline{41.6} & 11.8 & 41.4 & 30.7 & 18.0 & 7.9 & 17.7 & 7.1 & 6.1 & 31.2 & 36.3 & 20.8 & 19.9 & 7.4 & 6.6 & 9.0 & 5.2 & 34.2 & 14.8 & 18.7 & 24.2 & 27.3 & 19.1 & \textbf{45.3} & 37.5 & 31.6 \\
Waveform & 27.0 & \textbf{30.7} & 8.5 & 7.9 & 10.5 & 21.5 & 14.1 & 14.5 & 9.0 & 7.4 & 9.9 & 10.4 & 24.5 & 10.4 & 22.9 & 15.0 & 6.0 & 6.4 & 18.0 & 4.9 & \underline{28.3} & 12.2 & 9.4 & 17.5 & 5.7 & 6.0 & 10.4 & 7.0 & 12.6 \\
WBC & 85.0 & 58.6 & 95.2 & 73.1 & \textbf{98.7} & 72.3 & 89.5 & 86.8 & \underline{98.1} & \underline{98.1} & \underline{98.1} & 23.0 & 77.9 & 86.0 & 93.2 & 70.7 & 61.2 & 17.2 & 60.2 & 7.1 & 82.7 & 18.1 & 86.6 & 50.3 & 94.3 & 50.9 & 25.2 & 82.6 & 87.9 \\
WDBC & 96.5 & 98.3 & 98.3 & 98.1 & 98.2 & 98.0 & 97.4 & 95.8 & 91.2 & 77.3 & 97.7 & 20.2 & 95.3 & 97.4 & \underline{98.7} & 97.7 & 19.5 & 11.3 & 47.1 & 4.0 & 97.1 & 12.6 & 98.3 & 88.3 & 85.6 & 50.8 & 65.2 & \textbf{99.4} & 95.5 \\
Wilt & 12.2 & 15.7 & 8.6 & \underline{27.6} & 8.9 & 7.8 & 7.6 & 8.4 & 7.9 & 7.7 & 6.9 & 7.5 & 13.9 & 11.2 & 8.1 & 23.5 & 11.6 & 9.6 & 10.7 & 9.5 & 11.4 & 8.1 & 12.1 & \textbf{30.1} & 14.5 & 10.2 & 25.2 & 11.0 & 20.7 \\
wine & 13.6 & 12.5 & 13.1 & 11.5 & 15.4 & 16.6 & 18.1 & 15.9 & 13.1 & 35.8 & \textbf{54.5} & 15.0 & 12.2 & 13.0 & 13.3 & 12.1 & 15.7 & 17.5 & 31.6 & 9.1 & 14.3 & 25.1 & 12.9 & \underline{37.3} & 13.5 & 16.0 & 12.3 & 10.6 & 13.1 \\
WPBC & 40.2 & 38.6 & 39.4 & 37.1 & 37.7 & 38.4 & 38.9 & 40.3 & 37.7 & 35.2 & 37.8 & 36.0 & 39.7 & 39.1 & 39.1 & 37.8 & \textbf{45.4} & 39.7 & 39.2 & \underline{41.9} & 41.3 & 37.0 & 37.9 & 39.2 & 40.4 & 39.6 & 39.5 & 37.5 & 39.9 \\
\midrule
Average & 57.0 & 50.8 & 46.9 & 51.7 & 47.1 & 53.1 & 51.9 & 50.8 & 44.3 & 46.4 & 44.9 & 31.4 & \underline{57.4} & 53.1 & 53.6 & 51.2 & 34.7 & 22.0 & 41.5 & 27.3 & \textbf{59.8} & 46.5 & 46.9 & 53.3 & 49.6 & 44.0 & 52.5 & 56.3 & \underline{57.4} \\
Avg. rank & 9.74 & 13.15 & 17.00 & 13.74 & 15.93 & 11.12 & 13.15 & 13.25 & 17.05 & 17.05 & 17.65 & 21.30 & 9.50 & 12.23 & 12.08 & 14.65 & 21.35 & 24.45 & 19.11 & 23.18 & \underline{8.63} & 15.02 & 17.39 & 12.05 & 14.09 & 18.61 & 12.63 & 11.68 & \textbf{8.20} \\
\bottomrule
\end{tabular}}
\end{table*}

\begin{figure}[!t]
\centering
\includegraphics[width=0.49\textwidth]{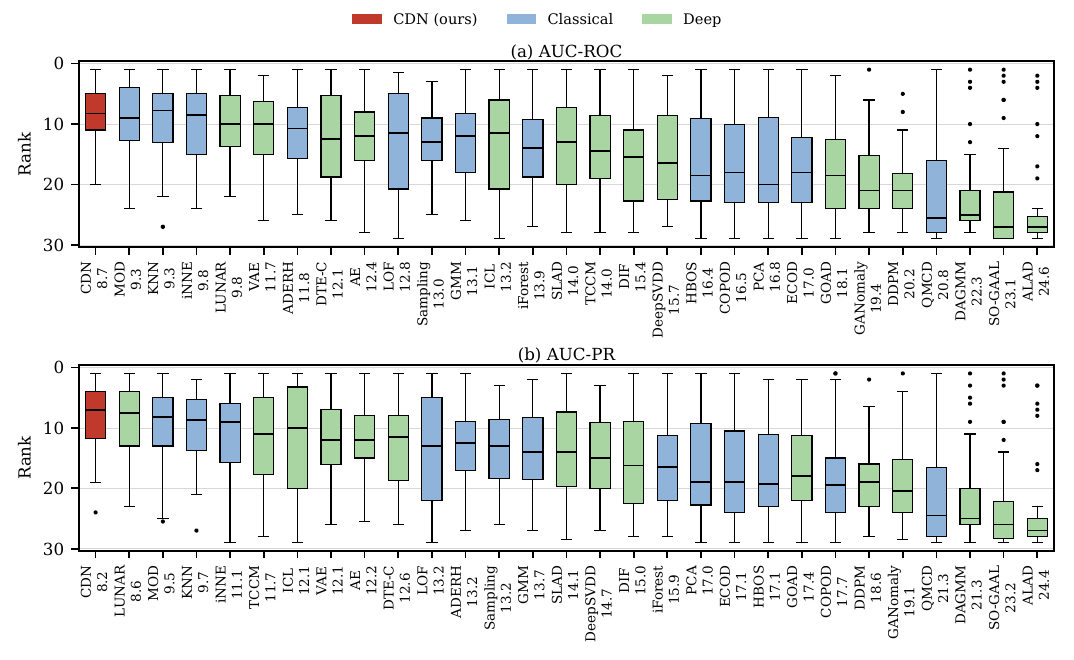}
\caption{Rank distributions over 46 datasets (lower is better). Methods are sorted by mean rank, shown below each name.}
\label{fig:rank}
\end{figure}

Tables~\ref{tab:roc} and~\ref{tab:pr} report the results on each dataset, and Fig.~\ref{fig:rank} shows the distribution of ranks. \ours\ has the best mean rank under both metrics and the highest average AUC-ROC, followed by a group of neighbourhood-based detectors, namely MOD, KNN, iNNE and LUNAR. Under AUC-PR, its average equals that of MOD and exceeds that of KNN but is below that of LUNAR. This gap comes from optdigits, smtp and glass, and above all from optdigits; without these three datasets the two averages differ by less than 0.1 points. Since the mean rank weighs every dataset equally, it is the more reliable summary here. Unlike KNN, MOD, and LUNAR, \ours\ neither retains the training set nor performs neighbour searches at inference time. \ours\ also clearly improves on AE and VAE, the reconstruction-based detectors that, like \ours, are data-free at inference. The ablation in Section~\ref{sec:ablation} isolates the contribution of the neighbour-mean target.

MOD is a particularly relevant comparator because it evaluates nearest-neighbour mean-shift displacements nonparametrically. However, the evaluated MOD configuration uses three shift iterations, whereas \ours\ is trained on one-step nearest-neighbour means. Thus, this comparison evaluates related detection strategies rather than exact regression of MOD's final multi-step output. \ours\ has a better mean rank than MOD under both metrics and the same or a higher average, while replacing its per-query neighbour searches with one forward pass of a small network: the displacement field that MOD recomputes for every query can be learned once and evaluated without the training data.

\subsection{Ablation Study}
\label{sec:ablation}
\begin{table}[t]
\centering
\caption{Ablation of \ours\ on 46 datasets. Avg.: average ($\times 100$); Rank: mean rank among the five variants.}
\label{tab:ablation}
\resizebox{0.48\textwidth}{!}{
\begin{tabular}{lcccc}
\toprule
 & \multicolumn{2}{c}{AUC-ROC} & \multicolumn{2}{c}{AUC-PR}\\
\cmidrule(lr){2-3}\cmidrule(lr){4-5}
Variant & Avg. & Rank & Avg. & Rank\\
\midrule
\ours & \textbf{81.2} & \textbf{2.49} & \textbf{57.4} & \textbf{2.39}\\
\midrule
ReLU activation & 80.3 & 2.74 & 53.9 & 3.04\\
$\tanh$ activation & 80.5 & 2.74 & 55.9 & 2.76\\
\midrule
Self-reconstruction target & 80.1 & 3.10 & 56.7 & 2.72\\
Skip connection $\bx+g_\theta(\bx)$ & 72.3 & 3.93 & 38.5 & 4.09\\
\bottomrule
\end{tabular}}
\end{table}

Table~\ref{tab:ablation} changes one component of \ours\ at a time, keeping the architecture, the training budget and all other settings fixed.

\textbf{Activation.} Replacing the Cauchy activation by ReLU or $\tanh$ lowers both metrics. The largest drop comes from ReLU under AUC-PR (3.5 points), concentrated on http, fraud, landsat, donors and cardio. This is consistent with the Remark~\ref{remark:2}: an unbounded ReLU network can let its prediction follow a query that lies far from the data. The bounded $\tanh$ activation is closer to \ours, in line with the observation that the guarantee of Proposition~\ref{prop:far} holds for bounded activations, and the Cauchy activation remains better under both metrics.

\textbf{Target.} Training the same network to reconstruct its input instead of the neighbour mean lowers the average AUC-ROC by 1.1 points and worsens the mean rank under both metrics. The neighbour-mean target therefore contributes beyond the bounded activation alone, as the two-dimensional example in Fig.~\ref{fig:toy} illustrates.

\textbf{Parameterisation.} Predicting the displacement through a skip connection, $f_\theta(\bx)=\bx+g_\theta(\bx)$ with anomaly score $\|g_\theta(\bx)\|$, causes the largest degradation: 8.9 points in AUC-ROC and 18.9 points in AUC-PR, with collapses on Lymphography, musk and WDBC. As explained in Section~\ref{sec:theory-far}, a bounded $g_\theta$ makes $f_\theta$ follow the query at large distances, preventing the residual score from growing without bound. These results demonstrate the importance of the direct parameterisation under our experimental configuration. They are also consistent with the far-field analysis: a bounded displacement network yields a bounded residual score, whereas the direct bounded-output network produces residual scores that diverge for sufficiently distant queries.

\subsection{Anomaly Types}
\label{sec:types}

\begin{figure}[!t]
\centering
\includegraphics[width=0.48\textwidth]{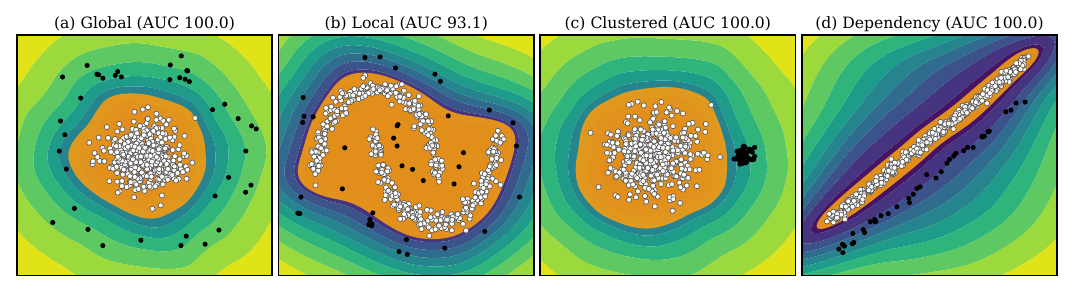}
\caption{\ours\ on the four types of anomalies of ADBench in two dimensions: (a) global, (b) local, (c) clustered, (d) dependency. Background: anomaly score (brighter is higher, as in Fig.~\ref{fig:toy}). Orange: region judged normal, with the threshold at the largest anomaly score of a normal test point. White: normal test points; black: anomalies. The AUC-ROC is computed on 500 normal test points and the anomalies shown.}
\label{fig:types}
\end{figure}

ADBench \cite{han2022adbench} distinguishes four types of anomalies. \emph{Global} anomalies lie far from all normal data. \emph{Local} anomalies lie close to the normal data but outside the region occupied by their neighbourhood. \emph{Clustered} anomalies form a compact group of their own. \emph{Dependency} anomalies take a normal value in every feature but violate the relation between features. Fig.~\ref{fig:types} shows one two-dimensional example of each type: a Gaussian cluster surrounded by scattered anomalies, two interleaving moons with anomalies outside and between them, a Gaussian cluster with a small group of anomalies next to it, and points on a line with anomalies on a parallel line. In these displayed examples, global and clustered anomalies receive larger scores than the normal observations. Proposition~\ref{prop:far} provides a theoretical explanation for the growth of residual scores at sufficiently large distances, but does not guarantee the detection of clustered anomalies near the normal-data region. Dependency anomalies are also separated, because the low-score region follows the direction of the data rather than the range of each feature. Local anomalies are detected when they lie outside the two moons but missed when they lie in the concave region between them, which gives the lowest AUC-ROC of the four (93.1). A point in this region has neighbours on both sides, their mean nearly coincides with the point, and its displacement is small, which is the situation described in Remark~\ref{rem:critical}. Anomalies in gaps of the normal data are therefore a limitation of \ours, shared with nonparametric mean shift.

% =================================================================

\subsection{Efficiency}
\begin{figure}[!htbp]
\centering
\includegraphics[width=0.48\textwidth]{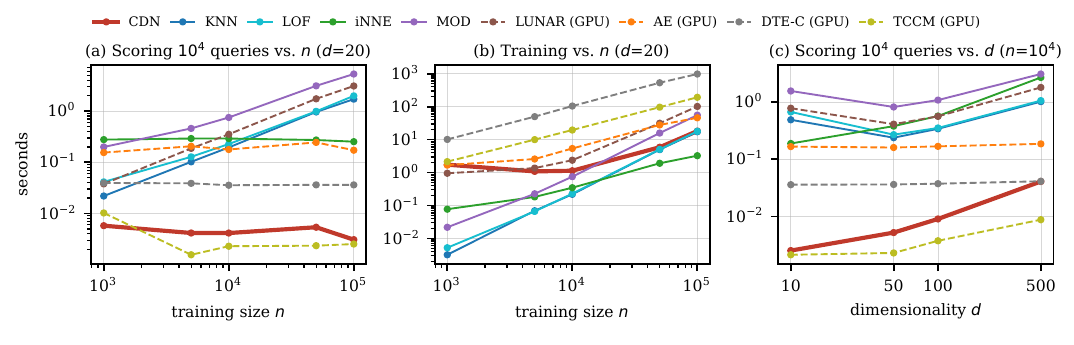}
\caption{Running time on synthetic Gaussian data (log scales). (a) Time to score $10^4$ queries as the training size $n$ grows with $d=20$. (b) Training time for the same settings. (c) Time to score $10^4$ queries as $d$ grows with $n=10^4$. CPU methods use a single thread; dashed lines are deep baselines run on the GPU.}
\label{fig:timing}
\end{figure}

We measure running time on synthetic Gaussian data, varying the training size $n$ from $10^3$ to $10^5$ with $d=20$ and the dimensionality $d$ from 10 to 500 with $n=10^4$, and scoring $10^4$ queries in every setting. All CPU computations, including those of \ours, use a single thread, so that the comparison with the classical detectors is not affected by parallelism. LUNAR, AE, DTE-C and TCCM run on the GPU with their default settings, which favours them over \ours. Fig.~\ref{fig:timing} shows the results.

\textbf{Inference.} The scoring time of \ours\ stays at 3 to 6 ms for $10^4$ queries over the whole range of $n$ (Fig.~\ref{fig:timing}(a)), because a query only passes through a network of fixed size. The scoring time of KNN, LOF, MOD and LUNAR grows with $n$, since every query is compared with the stored training set, and for LUNAR this holds even on the GPU. At $n=10^5$, \ours\ scores the queries about 560 times faster than KNN, 650 times faster than LOF, 1{,}000 times faster than LUNAR and 1{,}700 times faster than MOD, and about 80 times faster than iNNE, whose cost depends on its fixed sample size rather than on $n$. Among the other deep baselines, \ours\ on one CPU thread is faster than AE and DTE-C on the GPU and comparable to TCCM on the GPU. As $d$ grows (Fig.~\ref{fig:timing}(c)), the scoring time of \ours\ grows linearly as predicted by the $O(hd)$ cost and remains the lowest among the CPU methods, reaching 41 ms at $d=500$ compared with about 1 s for KNN and LOF. For $d=20$, \ours\ stores 5{,}652 parameters, about 22\,KB in single precision, whereas the neighbourhood-based detectors keep the $2\times10^6$ coordinates of a training set with $n=10^5$. A small, fixed memory footprint and millisecond scoring on a single CPU thread make \ours\ suitable for deployment where accelerators are unavailable or the training data cannot be retained, such as edge devices, embedded monitoring and on-premise systems that must score large volumes of records.

\textbf{Training.} Training \ours\ consists of the neighbour search for the targets and a fixed number of gradient steps. For small $n$ the fixed budget of 2{,}000 steps dominates and training takes about one second; for large $n$ the neighbour search dominates, and at $n=10^5$ \ours\ trains in 18 s, comparable to fitting KNN and LOF (17 to 18 s, which includes scoring the training set), and faster than MOD (55 s), AE (46 s), LUNAR (102 s), TCCM (193 s) and DTE-C (979 s). Only iNNE, which builds its model from small random samples, trains faster. Since the neighbour search is performed once and offline, it can be accelerated with approximate nearest-neighbour indices without affecting inference.

\begin{figure}
\centering
\includegraphics[width=0.48\textwidth]{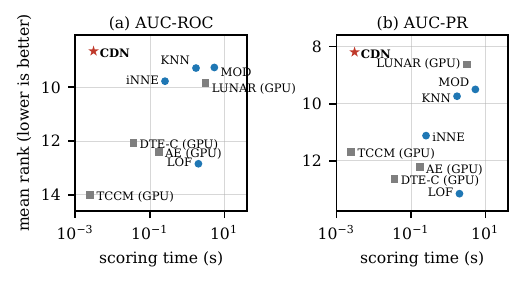}
\caption{Accuracy against inference cost. Vertical axis: mean rank over the 46 datasets among all 29 methods (lower is better). Horizontal axis: time to score $10^4$ queries with $n=10^5$ and $d=20$ (log scale). Squares are deep baselines run on the GPU.}
\label{fig:pareto}
\end{figure}

\textbf{Accuracy and cost.} Fig.~\ref{fig:pareto} combines the mean ranks of Section~\ref{sec:results} with the scoring times at $n=10^5$. Among the evaluated methods and implementation--device configurations, no baseline simultaneously achieves a lower reported mean rank and a shorter measured inference time than \ours. This empirical Pareto comparison is specific to the evaluated workloads and hardware settings. The detectors closest to \ours\ in mean rank, KNN, MOD and LUNAR under both metrics and iNNE under AUC-ROC, are 80 to 1{,}700 times slower, and the only detector with a comparable scoring time, TCCM on the GPU, has a mean rank of 14.0 under AUC-ROC and 11.7 under AUC-PR.
\subsection{Sensitivity Analysis}
\label{sec:sens}

\begin{table}
\centering
\caption{Sensitivity of \ours\ to its settings. Each row changes one setting from its default value, given in parentheses, and keeps the others at their defaults. Avg.: average over 46 datasets ($\times 100$); Rank: mean rank among the 29 methods; Pos.: position of \ours\ in that ranking.}
\label{tab:sens}
\small
\resizebox{0.4\textwidth}{!}{
\begin{tabular}{llcccccc}
\toprule
 & & \multicolumn{3}{c}{AUC-ROC} & \multicolumn{3}{c}{AUC-PR}\\
\cmidrule(lr){3-5}\cmidrule(lr){6-8}
Setting (default) & Value & Avg. & Rank & Pos. & Avg. & Rank & Pos.\\
\midrule
\multicolumn{2}{l}{Default} & 81.2 & 8.65 & 1 & 57.4 & 8.20 & 1 \\
\midrule
$h$ (128) & 32 & 80.6 & 9.32 & 3 & 55.7 & 9.39 & 3 \\
 & 64 & 81.0 & 8.93 & 1 & 56.6 & 8.61 & 2 \\
$\beta_0$ (0.2) & 0.01 & 81.1 & 8.55 & 1 & 57.2 & 8.39 & 1 \\
 & 0.1 & 81.2 & 8.65 & 1 & 57.2 & 8.11 & 1 \\
$T$ (2000) & 1000 & 81.2 & 8.72 & 1 & 56.8 & 8.46 & 1 \\
 & 4000 & 81.2 & 8.60 & 1 & 57.4 & 8.11 & 1 \\
$\gamma_0$ (0.6) & 0.3 & 81.0 & 9.05 & 1 & 57.2 & 8.80 & 2 \\
 & 1.0 & 81.1 & 8.74 & 1 & 57.0 & 8.32 & 1 \\
$k$ (5) & 3 & 81.1 & 8.95 & 1 & 57.2 & 8.70 & 2 \\
 & 10 & 80.8 & 9.21 & 2 & 57.1 & 8.66 & 2 \\
 & 20 & 80.5 & 9.16 & 2 & 56.8 & 8.92 & 2 \\
\bottomrule
\end{tabular}}
\end{table}

Table~\ref{tab:sens} changes one setting at a time from the default configuration, where $\beta_0$ and $\gamma_0$ denote the initial values of $\beta_j$ and $\gamma_j$. \ours\ is first or second in mean rank under both metrics in every setting except $h=32$. The average AUC-ROC varies by at most 0.7 points across all settings. The hidden width is the setting with the clearest effect: AUC-PR decreases from 57.4 to 56.6 with $h=64$ and to 55.7 with $h=32$. The initial activation parameters and the number of training steps change the average AUC-ROC by at most 0.2 points and the average AUC-PR by at most 0.6 points. Varying the neighbourhood size from $k=3$ to $k=20$ changes the average AUC-ROC by at most 0.7 points and the average AUC-PR by at most 0.6 points, and $k=5$ gives the best results under both metrics. The initial $\alpha_j$ had little effect in preliminary experiments and is fixed to $0.1$.

% \section{Conclusion}

\section{Conclusion}

We presented \ours, a compact anomaly detector that learns normal-data neighbourhood means with a single-hidden-layer Cauchy network and scores queries by prediction residuals. The supervision is related to mean shift and population density-score geometry.

For a structured family of Student-$t$-type expert densities, we established exact Cauchy score representation and a constructive approximation bound for the corresponding population neighbourhood-mean map. We further established conditional preservation of score margins and asymptotic far-field growth of the residual score.

Experiments on 46 tabular datasets demonstrated competitive detection accuracy and the best reported mean rank under both AUC-ROC and AUC-PR. CDN additionally avoids retaining the normal reference set and performing neighbour searches during inference, providing a favourable accuracy--efficiency trade-off.

Our method has several limitations. Population mean-shift scores can vanish at low-density stationary points, so local anomalies in gaps between normal clusters may be missed. The theoretical approximation results concern fixed-radius population maps rather than finite-sample KNN targets or the outcome of gradient-based training. CDN also assumes numerical inputs and normal-only training data. Future work includes mixed-type features, contaminated training sets, and adaptive neighbourhood supervision.

% \section*{Acknowledgment}
% Yang Cao is supported by the Guangdong Basic and Applied Basic Research Foundation (Grant No.~2025A1515110204). 

% Zihao Guo is supported by the China Postdoctoral Science Foundation(Certificate Number: 2026M793346 ). 

% The authors used LLM to draft and revise parts of the text, write the plotting code, and check the reported numbers. The authors reviewed all AI-generated content and take full responsibility for the paper.
% \clearpage
\bibliographystyle{IEEEtran}
\bibliography{ref}

\end{document}